%% file: main.tex
\documentclass[10pt,a4paper]{article}
\usepackage{consistency}
\input{authors}
\hypersetup{pdftitle={A Consistency Principle for Federated Learning},pdfauthor={\manuscriptpdfauthors}}
\begin{document}
\paperheading{A Consistency Principle for Federated Learning}{}
\begin{paperabstract}
Two clients can start from the same model and still contribute updates that, once averaged, point away from a stationary point of the global loss. This observation motivates a broader view of consistency in federated learning. We call a local computation consistent with a specified global operation to the extent that it preserves the update, objective or constraint that the operation is meant to implement. Drawing on the related literature, we use this principle to examine client trajectories, adaptive directions, sharpness perturbations, primal--dual updates and decentralized mixing. A common notation distinguishes bias in an aggregated direction from sampling variance, and relates each case to the stationarity, stability or margin bound available for it. A controlled quadratic experiment isolates the bias introduced by local scaling. Published vision experiments illustrate both improvements and settings in which a smaller discrepancy accompanies lower accuracy. These examples suggest which client states to retain, which deviations to measure and how to weigh a correction against its communication cost.
\end{paperabstract}
\input{sections/01-introduction}
\input{sections/02-principle}
\input{sections/03-trajectories}
\input{sections/04-sharpness}
\input{sections/05-duals}
\input{sections/06-topology}
\input{sections/06a-synthesis}
\input{sections/07-evidence}
\input{sections/08-deployment}
\input{sections/09-discussion}
\FloatBarrier
\begingroup
\renewcommand{\bibsection}{\section*{References}\addcontentsline{toc}{section}{References}}
\renewcommand{\bibfont}{\normalfont\small}
\bibliographystyle{ieeetr}
\bibliography{references}
\endgroup
\clearpage
\input{sections/appendix}

\end{document}

%% file: authors.tex
\newcommand{\manuscriptpdfauthors}{Yan Sun, Dacheng Tao}

\newcommand{\manuscriptauthors}{%
  Yan Sun\authorref{1},\quad Dacheng Tao\authorref{2}%
}

\newcommand{\manuscriptauthornotes}{%
  \authornote{1}{The University of Sydney, Sydney, Australia.
    Email: \nolinkurl{ysun9899@uni.sydney.edu.au}.}%
  \authornote{2}{Nanyang Technological University, Singapore.
    Email: \nolinkurl{dacheng.tao@ntu.edu.sg}.}%
}

%% file: sections/01-introduction.tex
\section{Beyond client drift}\label{sec:introduction}
Federated averaging lets a hospital, a mobile device or a data silo take several optimization steps before sending a single update. This exchange saves communication and leaves training examples where they were collected \cite{mcmahan2017communication,kairouz2021advances,li2020review}. It also changes the computation. Each client evaluates later gradients on a trajectory shaped by its own data. A server that averages these trajectories recovers a joint update only under particular conditions. The familiar name for one consequence is \emph{client drift} \cite{karimireddy2020scaffold}. Even clients that start each round at identical weights can drift apart before the round ends.

Reliable aggregation requires client contributions to be defined on a consistent basis. When clients apply different local transformations or rely on different reference points, simply averaging their contributions may no longer preserve the intended global objective. Similar issues arise when clients independently maximize the loss inside a perturbation ball, keep a multiplier fixed through a long absence, or substitute a neighborhood average for a population average. Each case calls for a different reference against which to judge agreement.

This article revisits the \emph{consistency principle} developed in a doctoral thesis on federated learning \cite{sunthesis}. We use it to ask what global operation a client update is meant to preserve and how a departure from that operation might affect learning. Local computation is valuable precisely because it can differ across clients. At a global optimum the client gradients can be large and cancel, and the corresponding client multipliers may legitimately differ. An architecture that forced every local quantity to coincide could destroy that solution. What matters is whether the local quantities, after the operations that actually use them, still implement the intended global update, objective or constraint. We give this distinction a mathematical form in Section~\ref{sec:principle}.

\lead{Where the principle fits} Early analyses of distributed optimization established the computation--communication trade-off for mini-batches, periodic averaging, and local SGD \cite{li2014efficient,gower2019sgd,yu2019parallel,stich2018local,woodworth2020local,woodworth2020minibatch,gorbunov2021local}. The choice of step size and local horizon already matters under full participation \cite{Lin2020Dont,dieuleveut2019communication,spiridonoff2021communication}. FedNova identifies a specific objective inconsistency when unequal local work changes the effective client weights \cite{wang2020fednova}. SCAFFOLD corrects gradient drift with control variates \cite{karimireddy2020scaffold}. Mime asks how to reproduce a centralized optimizer with federated data access \cite{karimireddy2020mime}. Work on local adaptivity directly connects scaling to convergence to a mismatched objective \cite{wang2021local}. These precedents motivate using a reference operation. Choosing an inner maximization, a dual constraint or a mixing step as that reference lets us ask the same question in other parts of a training round.

The setting also differs from a tightly coupled cluster. Distributed training surveys organize systems by synchronization and consolidation \cite{peteiro2013survey,verbraeken2020survey,khan2023decentralized}. In FL the identity, frequency and duration of a client's return determine whether a historical state can be used at all. This systems distinction becomes part of the mathematical reference for relaxed initialization and dual updates.

\lead{Scope of the evidence} The thesis draws on collaborative studies of relaxed initialization \cite{sun2023understanding}, local adaptivity \cite{sun2023efficient}, sharpness-aware training \cite{sun2023dynamic}, primal--dual alignment \cite{sun2024fedpd} and communication topology \cite{sun2025towards}. We use these studies as worked examples alongside related approaches. Theorems~\ref{thm:init}--\ref{thm:network} summarize the results of the respective studies in common notation. Section~\ref{sec:synthesis} compares what these guarantees control. Appendix~\ref{sec:derivations} develops the underlying proofs, with the source estimates and their conditions stated explicitly. The scope is a shared prediction model trained with synchronous rounds, together with the corresponding fixed-graph decentralized protocol. Personalized objectives, directed graphs and privacy mechanisms change the reference or the information available to estimate it. We discuss these boundaries where they affect deployment.

%% file: sections/02-principle.tex
\section{A reference for federated computation}\label{sec:principle}
\subsection{Objective, sampling and units of analysis}
Client $i\in[m]$ holds $n_i$ independent examples $z_{ij}$ from its own distribution $\mathcal D_i$. For weights $p_i\geq0$ summing to one, define the empirical and population objectives by
\begin{equation}\label{eq:risk}
 f(\w)=\sum_i p_i f_i(\w),\qquad f_i(\w)=\frac1{n_i}\sum_j\ell(\w;z_{ij}),\qquad F(\w)=\sum_i p_i\E_{z\sim\mathcal D_i}\ell(\w;z).
\end{equation}
Weights proportional to $n_i$ give each example equal influence. Uniform weights give each client equal influence. All five case studies below use the uniform-client convention in their stated results. In round $t$ the server samples $\cP_t$ of size $m_p$ from $m$ clients. Each takes $K$ local steps before aggregation, for $T$ rounds in total. A uniform active-client mean is unbiased for a fixed population of equally weighted messages under uniform sampling without replacement. Unequal inclusion probabilities $\pi_i>0$ instead give the unbiased sum $\sum_{i\in\cP_t}p_i u_i/\pi_i$ for fixed messages $u_i$. Model-dependent availability and unequal local work require their own analysis \cite{wang2020fednova,jhunjhunwala2022fedvarp}. We use $\E_t$ for expectation conditional on the history before cohort selection in round $t$.

\begin{definition}[Reference and consistency residual]\label{def:reference}
Specify a global target, a point in the training round, and an operation with output in a normed space. Holding the available history $H_t$ fixed, let $R_t$ be the output of the corresponding globally informed operation. Let $U_t=\mathcal A_t(s_{1,t},\ldots,s_{m,t};\xi_t)$ be the output implemented from local states $s_{i,t}$, where $\xi_t$ includes sampling and local randomness. Relative to a history-measurable $R_t$, the \emph{systematic residual} is $b_t=\E_t U_t-R_t$, and the \emph{mean-square residual} is $D_t=\E_t\norm{U_t-R_t}^2$. We call the operation consistent in conditional mean if $b_t=0$, and mean-square consistent along a sequence if $\E D_t\to0$. For a deterministic constraint, $R_t$ may be its required residual, such as zero in the consensus subspace.
\end{definition}

\noindent For Euclidean outputs, with $v_t=\E_t\norm{U_t-\E_tU_t}^2$, the identity $D_t=\norm{b_t}^2+v_t$ is exact. A changing, random oracle can also be compared pathwise by coupling its randomness to that of the local implementation. To interpret a diagnostic one must name this coupling or fix the reference in advance. A single client's empirical distance from a sampled mean is a different quantity from bias relative to the population. A dispersion measure can be zero even when all messages share the same bias.

\principlepanel{\begin{definition}[Consistency principle]\label{def:principle}
Federated training should preserve, to the accuracy required by the learning target, the global operation that will consume the local state. To apply the principle, specify the state, the operation, its globally compatible reference and the discrepancy by which they are compared. A correction is justified when this discrepancy enters a bound on the target error, or when a controlled intervention links its reduction to the target metric at an acceptable resource cost. The reference may specify a joint operation, a set of compatible client states, or a projection onto a constraint. For a set-valued reference $\mathcal R_t$, use $\operatorname{dist}(s_t,\mathcal R_t)=\inf_{r\in\mathcal R_t}\norm{s_t-r}$.
\end{definition}}

The target determines the resolution of the comparison. Model disagreement concerns client endpoints. Adaptive bias concerns the \emph{aggregated preconditioned direction}. Sharpness concerns the order of averaging and maximization. At a differentiable consensus optimum of $\sum_i p_i f_i(\w_i)$ with constraints $\w_i=\w$, a compatible multiplier in the weighted Lagrangian satisfies $\lam_i=-\nabla f_i(\w)$ and $\sum_i p_i\lam_i=0$. Those multipliers can differ from every other client's. Table~\ref{tab:map} identifies one practically accessible quantity for each comparison.

\begin{table}[tb]
\caption{Five interpretations of consistency. Each observable has a specified reference point in the training round. Proxy measurements inherit the sampling and staleness of the messages used to compute them.}\label{tab:map}
\centering\small\renewcommand{\arraystretch}{1.17}
\begin{tabularx}{\linewidth}{@{}p{.145\linewidth}Y Y@{}}\toprule
State & Globally compatible reference & Observable or computable proxy \\ \midrule
Local model & Endpoints whose combined displacement tracks descent on $f$ & Mean squared endpoint distance to the current aggregate, tagged with cohort and time.\\
Adaptive update & A globally meaningful preconditioned direction & Mean and spread of normalized local updates relative to the previous global displacement.\\
Perturbation & One admissible ascent direction for $f$ & Local-to-shared perturbation distances, or the first-order objective gap in \eqref{eq:sharp-gap}.\\
Multiplier & Per-client stationarity and aggregate dual balance & Local subproblem residual on returning clients, stratified by time since last participation.\\
Mixed model & Population consensus projection of the client matrix & Disagreement after gossip, or a spectral upper bound from the deployed graph.\\\bottomrule
\end{tabularx}
\end{table}

\subsection{When a residual has an optimization or statistical price}
\begin{proposition}[Descent with a mismatched mean direction]\label{prop:descent}
Let $f$ be $L$-smooth and bounded below by $f_\mathrm{lo}$. Suppose $\w^{t+1}=\w^t-hU_t$ with finite conditional second moments and $0<h\leq1/L$. Use $R_t=\nabla f(\w^t)$ in Definition~\ref{def:reference}. Then
\begin{equation}\label{eq:descent}
 \frac1T\sum_{t=0}^{T-1}\E\norm{\nabla f(\w^t)}^2
 \leq \frac{2(f(\w^0)-f_\mathrm{lo})}{hT}
 +\frac1T\sum_{t=0}^{T-1}\E\norm{b_t}^2+\frac{Lh}{T}\sum_{t=0}^{T-1}\E v_t .
\end{equation}
\end{proposition}
Lemma~\ref{lem:descent} gives the one-step inequality behind this result.

The variance term carries a factor $h$. A persistent bias can survive decreasing step sizes or increasing batch size. The proposition applies directly when the desired direction is the gradient. Preconditioned descent can instead be studied in an appropriate metric or on an auxiliary sequence. A primal--dual solver uses stationarity and feasibility together. Applying \eqref{eq:descent} to their raw messages would conflate an intentional change of geometry with an error.

For statistical performance let $A(S)$ be the trained model and let $\w_F^\star$ minimize $F$. Adding and subtracting the empirical risk and using $\E f(\w_F^\star)=F(\w_F^\star)$ gives
\begin{equation}\label{eq:excess}
 \E[F(A(S))-F(\w_F^\star)]\leq
 \underbrace{\E[F(A(S))-f(A(S))]}_{\text{expected generalization gap}}+
 \underbrace{\E[f(A(S))-\inf_{\w} f(\w)]}_{\text{empirical suboptimality}} .
\end{equation}
Two runs on samples differing in one example, coupled with the same algorithmic randomness, have uniform stability $\varepsilon$ if the expected absolute change in test loss is at most $\varepsilon$ for every replacement and test point. This controls the signed expected gap above by the replace-one argument \cite{hardt2016train,kuzborskij2018data,zhang2022stability}. A stable procedure can still underfit. Conversely, a small gradient on a general nonconvex objective establishes stationarity and needs additional geometry to control empirical suboptimality. These distinctions determine which conclusion each of the following analyses supports.

%% file: sections/03-trajectories.tex
\section{Correcting the federated update}\label{sec:update}
\subsection{Model trajectories and relaxed initialization}\label{sec:init}
With a common start and no extrapolation, the normalized local-SGD displacement is an average of gradients evaluated at different points. Define the full-client trajectory as the one client $i$ would have followed if selected. Under unbiased stochastic gradients and uniform sampling independent of these potential trajectories, its conditional mean differs from $\nabla f(\w^t)$ by
\begin{equation}\label{eq:path}
 b_t=\frac1K\sum_{k=0}^{K-1}\sum_i p_i\E_t[\nabla f_i(\w_{i,k}^t)-\nabla f_i(\w^t)],\qquad
 \norm{b_t}^2\leq\frac{L^2}{K}\sum_{k,i}p_i\E_t\norm{\w_{i,k}^t-\w^t}^2.
\end{equation}
The inequality follows by Jensen's inequality and local smoothness. It concerns the whole local path. The endpoint alone is cheaper to observe and can miss an excursion that returns to the same point. Proximal anchoring \cite{li2020federated,jiang2024stabilized}, control variates \cite{karimireddy2020scaffold} and drift decoupling \cite{gao2022feddc} intervene on different terms along this path.

\method{FedInit} changes the initial condition \cite{sun2023understanding,shen2026rethinking}. Let $h_i^t$ denote client $i$'s last endpoint, seeded at $h_i^0=\w^0$ and retained while the client is inactive. For an active client set
\begin{equation}\label{eq:ri}
 \w_{i,0}^t=\w^t+\beta(\w^t-h_i^t),\qquad
 \w_{i,k+1}^t=\w_{i,k}^t-\eta\g_{i,k}^t.
\end{equation}
The minus sign in front of $h_i^t-\w^t$ matters. It gives a client that repeatedly drifts in one direction room to move before averaging. It also introduces an initial displacement into the actual uploaded update. Equation~\eqref{eq:path} alone therefore does not bound the full \method{FedInit} update. A client keeps one model vector between participations. The server can keep its ordinary averaging rule and transmit the same model-sized messages as in \method{FedAvg}.

\lead{Why an inconsistent start can help} Consider two equal-weight quadratics $f_i(w)=(w-a_i)^2/2$ with $a_i\in\{-1,1\}$, global start $w=0$ and previous endpoint $d a_i$, $d>0$. After $K$ gradient steps write $s=(1-\eta)^K$ with $0<\eta<1$. The new endpoint is $[(1-s)-s\beta d]a_i$. The root-mean-square distance from the mean is $\beta d$ at initialization and $|(1-s)-s\beta d|$ at upload. Thus any $0<\beta<2(1-s)/(sd)$ improves endpoint agreement in this example. The optimum depends on the local horizon and the history. Symmetry keeps the average at zero even without the correction, a reminder that a reduced dispersion can leave the aggregate unchanged. Figure~\ref{fig:analytic} contrasts the two measurement times.

\begin{theorem}[Relaxed initialization, \cite{sun2023understanding}]\label{thm:init}
Suppose the losses are $L$-smooth, local gradients are unbiased with variance at most $\sigma_l^2$, and $\sum_i p_i\norm{\nabla f_i}^2\leq G_h^2+B^2\norm{\nabla f}^2$. Use the refined history estimate in Appendix~\ref{app:init}. Let $K\ge2$, $1<m_p<m$, $0\leq\beta<\sqrt{2}/12$ and $\eta\leq\min\{m_p/(2mKL),1/(m_pKL)\}$, with a positive descent coefficient as in the source analysis. For $D_0=f(\w^0)-\inf f$, the global iterates satisfy
\begin{equation}\label{eq:ri-bound}
 \frac1T\sum_{t<T}\E\norm{\nabla f(\w^t)}^2
 \ \lesssim_\beta\ \frac{D_0}{\eta KT}+\frac{\eta L}{m_p}\bigl(\sigma_l^2+K G_h^2\bigr),
\end{equation}
where $\lesssim_\beta$ suppresses the constants in the descent estimate.
\end{theorem}

Choosing $\eta$ of order $\sqrt{m_p/(KT)}$ where admissible gives the familiar inverse-square-root dependence, with $KG_h^2$ still in the numerator. Under the Polyak--\L{}ojasiewicz condition $2\mu(f-\inf f)\leq\norm{\nabla f}^2$, the published last-iterate objective-gap rate is $\widetilde{\bigO}((D_0+L(\sigma_l^2+KG_h^2))/(m_pKT))$ for the corresponding horizon-dependent step \cite{sun2023understanding}. For nonzero relaxation, Appendix~\ref{app:init} retains the weighted history term that must be controlled for this simplification.

To see where stability enters, let $\Delta_t=m^{-1}\sum_i\E\norm{h_i^t-\w^t}^2$. The original analysis bounds part of the generalization gap by a multiple of the time average of $\sqrt{\Delta_t}$ \cite{sun2023understanding}. A subsequent analysis assumes equal client sizes $n$, bounded Lipschitz losses and a round-wise step $c/(t+1)$. Its balanced stability bound contains a factor $(1+2\beta)^{-1/(1+cKL)}$ relative to its $\beta=0$ expression \cite{shen2026rethinking}. This finite-horizon comparison also requires the schedule compatibility in Appendix~\ref{app:init}. For a fixed positive $\beta$ the condition can fail late in training. The choice of $\beta$ therefore depends on which quantity is monitored and when.

Gradient corrections supply other ways to act on \eqref{eq:path}. Classical control-variate and recursive estimators suppress noise in finite sums \cite{johnson2013accelerating,defazio2014saga,nguyen2017sarah,fang2018spider}. In FL, variance reduction can target within-client sampling, active-cohort variation, or discrepancy between client models \cite{li2023effectiveness,wang2023communication,zhang2025federated}. Historical momentum also has a location. Server and slow momentum use previous aggregate displacements \cite{Wang2020SlowMo:,sun2024role}, whereas client momentum and drift control enter each local trajectory \cite{liu2020accelerating,xu2021fedcm,ozfatura2021fedadc,liu2023enhance}. Two-sided and adaptive momentum couple the two levels \cite{khanduri2021stem,das2022faster,cheng2024momentum,ba2024fast}. For a client that rarely returns, storing history centrally changes both its age and the cost of retrieving it.

\subsection{Adaptive directions and a common amendment}\label{sec:lada}
Adaptive methods use gradient history to rescale coordinates \cite{duchi2011adaptive,tieleman2012lecture,kingma2014adam,zaheer2018adaptive}. Applying the scaling to an aggregated displacement gives server-side adaptivity \cite{reddi2021adaptive,wang2022communication}. Applying it on each client changes the messages that reach the server \cite{wang2021local,mukherjee2023locally}. At a fixed point let $P_i$ be a client's diagonal preconditioner, $g_i$ its direction, and let bars denote weighted means. The identity
\begin{equation}\label{eq:pcov}
 \sum_i p_i P_i g_i-\bar P\bar g=\sum_i p_i(P_i-\bar P)(g_i-\bar g)
\end{equation}
identifies a covariance term. A positive diagonal $\bar P$ preserves stationarity of $f$ when applied to its gradient. The left-hand aggregate can move even at $\bar g=0$, as in the opening example. If the desired centralized preconditioner is $P_\star$, $(\bar P-P_\star)\bar g$ is a further mismatch. A global second moment formed from global gradients generally differs from an average of local second moments.

\method{FedLADA} \cite{sun2023efficient} mixes a local adaptive direction with a common historical direction. For one client step, write $g=\g_{i,k}^t$. With moment coefficients $0\leq\beta_1,\beta_2<1$, update the first and second moments by
\begin{equation}\label{eq:lada-moments}
 m^+=\beta_1m+(1-\beta_1)g,\quad v^+=\beta_2v+(1-\beta_2)g^{\odot 2},\quad
 \hat v^+=\max(\hat v,v^+),\quad a_{i,k}^t=m^+\oslash\sqrt{\hat v^+}.
\end{equation}
All operations on coordinates are elementwise. Each round starts with $m=0$ and $v=\hat v$ equal to the broadcast second moment. The server seeds this estimate at $\epsilon_v^2\mathbf1$ and averages the active terminal $\hat v$ values. With local and server steps $\eta_l,\eta_g$, the client uses
\begin{equation}\label{eq:lada}
 d_{i,k}^t=\alpha a_{i,k}^t+(1-\alpha)q^t,\qquad
 q^t=\frac{\w^{t-1}-\w^t}{K\eta_l\eta_g},\qquad \w_{i,k+1}^t=\w_{i,k}^t-\eta_l d_{i,k}^t .
\end{equation}
Here $0<\alpha\leq1$, $q^0=0$, and the denominator uses the steps that generated the previous displacement. Within a fixed step the spread satisfies $\sum_i p_i\norm{d_i-\bar d}^2=\alpha^2\sum_i p_i\norm{a_i-\bar a}^2$. The mean error relative to a desired direction $r$ is $\alpha(\bar a-r)+(1-\alpha)(q^t-r)$. These identities explain opposite failure modes. A large $\alpha$ retains heterogeneous scaling. A very small $\alpha$ copies an old direction even when the gradient has rotated.

\begin{theorem}[Local adaptive directions, \cite{sun2023efficient}]\label{thm:lada}
Set $\mathbf y^t=(\w^t-(1-\alpha)\w^{t-1})/\alpha$ with $\w^{-1}=\w^0$. Assume smoothness, bounded coordinate-wise gradients, bounded stochastic variance and bounded gradient dissimilarity. The second moment has a positive floor $\epsilon_v^2$. Under the cited step-size restrictions, including $K>\beta_1/(1-\beta_1)$ for the stated descent coefficient, and with $K\eta_g\eta_l\asymp\sqrt{m_pK/T}$ and $\eta_l\asymp1/\sqrt{KT}$, the auxiliary iterates satisfy
\begin{equation}\label{eq:lada-rate}
 \frac1T\sum_{t=0}^{T-1}\E\norm{\nabla f(\mathbf y^t)}^2
 \;=\;\bigO\!\left((m_pKT)^{-1/2}+(KT)^{-1}+T^{-1}+T^{-3/2}\right).
\end{equation}
\end{theorem}

The leading term describes stationarity along $\mathbf y^t$ for fixed admissible hyperparameters. Uploading the second moment, broadcasting the amendment and tuning $\alpha$ all matter for elapsed time. Client-specific rates, restarted dual averaging, automatic tuning and memory-efficient optimizers occupy different points in that trade-off \cite{sun2023fedlalr,li2023fedda,kim2024adaptive,wu2023faster,lee2026efficient,zhou2025fedcada}.

A separate systems question is how much of the direction to transmit. Sign-based update rules, sparse aligned updates and federated AdamW explicitly change this cost \cite{tang2024fedlion,deng2025towards,liu2026fedadamw}. Their comparison depends on the total number and precision of transmitted coordinates as well as the number of rounds.

%% file: sections/04-sharpness.tex
\section{Whose loss should the perturbation sharpen?}\label{sec:sharp}
Sharpness-aware minimization (SAM) measures loss after an adversarial weight perturbation of radius $\rho$ \cite{hochreiter1994simplifying,foret2021sharpnessaware}. This construction has two distinct federated realizations. With $B_\rho=\{\epsv:\norm{\epsv}\leq\rho\}$, define
\begin{equation}\label{eq:sharp-obj}
 \Phi_{\rm sh}(\w)=\max_{\epsv\in B_\rho}\sum_i p_i f_i(\w+\epsv),\qquad
 \Phi_{\rm loc}(\w)=\sum_i p_i\max_{\epsv_i\in B_\rho} f_i(\w+\epsv_i).
\end{equation}
The first permits one perturbation for the deployed global model. The second lets each data distribution choose its own. The product feasible set contains the diagonal, so $\Phi_{\rm sh}\leq\Phi_{\rm loc}$. When the maxima exist, equality holds exactly when the positive-weight clients have a common maximizing perturbation. The appropriate objective depends on whether the predictor must tolerate a shared weight change or distinct changes on different clients. The distinction matters as soon as shared-model flatness is the stated goal.

\begin{proposition}[First-order cost of independent perturbations]\label{prop:sharp}
If each $f_i$ has $L$-Lipschitz gradient on $\w+B_\rho$, then
\begin{equation}\label{eq:sharp-gap}
 \left|\Phi_{\rm loc}(\w)-\Phi_{\rm sh}(\w)
 -\rho\left(\sum_i p_i\norm{\nabla f_i(\w)}-\norm{\nabla f(\w)}\right)\right|\leq L\rho^2 .
\end{equation}
\end{proposition}
See Lemma~\ref{lem:sharp-linear} for the uniform Taylor bound.

The first-order gap is zero for aligned nonzero gradients and can be large when large gradients cancel. For the two quadratics of Section~\ref{sec:init}, the shared objective at $w=0$ is $(1+\rho^2)/2$ and the separate objective is $(1+\rho)^2/2$. The gap is exactly $\rho$ even though both clients evaluate the same model. Both objectives in this symmetric example attain their minimum at the origin. The gap identifies a change of criterion and need not move its minimizer.

\begin{figure}[tb]
\centering\includegraphics[width=\linewidth]{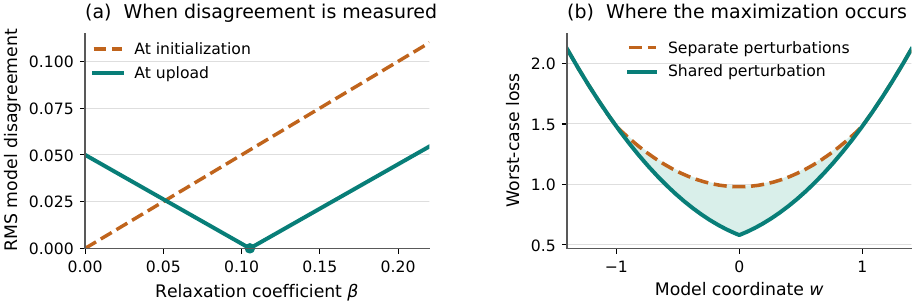}
\caption{Two clients with local minima at $-1$ and $+1$. Left, a larger initial separation can lead to closer endpoints after local training ($d=0.5$, $K=1$, $\eta=0.05$). Right, separate worst-case perturbations raise the loss above the shared worst case near the point of gradient cancellation ($\rho=0.4$). Both panels plot closed-form expressions.}\label{fig:analytic}
\end{figure}

\lead{Correcting the inner and outer problems} Local SAM \cite{qu2022generalized,caldarola2022improving} uses a normalized client gradient in the inner step. \method{FedSMOO} maintains a server perturbation $\epsv^t$, a client perturbation multiplier $\bm\mu_i$ and a separate model multiplier \cite{sun2023dynamic}. In the thesis notation, with $N_\rho(v)=\rho v/\norm v$ for $v\ne0$ and a zero-vector convention at $v=0$, a client forms
\begin{equation}\label{eq:smoo}
 \widehat{\epsv}_i=N_\rho\!\big(a(\g_i-\bm\mu_i)-\epsv^t\big),\qquad
 \bm\mu_i\gets\bm\mu_i+(\widehat{\epsv}_i-\epsv^t)/a .
\end{equation}
Here $a>0$ sets the perturbation calibration scale. The client evaluates a second gradient at $\w_i+\widehat{\epsv}_i$ and adds a dynamic model correction. The server normalizes the mean of the returned $a\bm\mu_i-\widehat{\epsv}_i$ to form its next reference. At a fixed model, replacing a shared perturbation by a local one changes the corresponding local gradient by at most $L\norm{\widehat{\epsv}_i-\epsv^t}$. This supplies a mechanism for calibration to matter. The server direction itself is an estimate of a globally useful perturbation. FedGAMMA and subsequent approaches estimate global directions with other information and communication budgets \cite{dai2023fedgamma,fan2024locally,caldarola2025beyond}.

The radius and the inner solver are separate design choices. Scale-adaptive neighborhoods change the perturbation set \cite{kwon2021asam}. Surrogate-gap minimization, gradient-norm penalties and sparse perturbations change the local approximation or its cost \cite{zhuang2022surrogate,zhao2022penalizing,mi2022make}. These choices modify the corresponding reference in \eqref{eq:sharp-obj}. A comparison that changes both the radius convention and the degree of cross-client alignment cannot isolate the latter.

\begin{theorem}[Shared perturbations, \cite{sun2023dynamic}]\label{thm:smoo}
Suppose the local subproblems are solved exactly, the $f_i$ are $L$-smooth, and the virtual client history satisfies the gradient-tracking initialization of the analysis. Let $b=\beta_s>0$ be small enough for the positive coefficients in Appendix~\ref{app:sharp}, and let $z^t=\bar\w^t$ be the active mean before model-dual correction. For initial gap $D_1=f(z^1)-\inf f$, stored-model disagreement $\delta_0$, and $p,q,\zeta>0$ specified there, the potential estimate in \cite{sun2023dynamic,sunthesis} gives
\begin{equation}\label{eq:smoo-rate}
 \frac1T\sum_{t=1}^T\E\norm{\nabla f(z^t)}^2
 \leq\frac{D_1+q\delta_0}{b\zeta T}
       +\frac{(1+30pb)L^2\rho^2}{\zeta}.
\end{equation}
Taking $\rho=O(1/\sqrt{m_pT})$ gives the reported $O(1/T)$ order for fixed admissible parameters.
\end{theorem}

The gradient in \eqref{eq:smoo-rate} is that of the empirical $f$. The radius shrinks with the horizon. Practical finite-$K$ runs use approximate local solutions. A fixed-radius guarantee for a robust objective would require a separate argument.

For a depth-$D$ bias-free network with 1-Lipschitz activations vanishing at zero, and layer-size bound $d$, let $\norm z\leq R$ and use $n$ examples per client. Write $V(W)=\prod_j\norm{W_j}_2^2\sum_j\norm{W_j}_F^2/\norm{W_j}_2^2$ for nonzero layers. Under the spectral normalization and sampling and fixed-prior conditions in Appendix~\ref{app:sharp}, for a normalized margin $0<\gamma\le1$, the spectral margin analysis \cite{neyshabur2018a,farnia2018generalizable,sun2023dynamic} bounds the population zero-margin error of the globally perturbed predictor, with probability $1-\delta$, by
\begin{equation}\label{eq:margin}
 G_0\ \leq\ \widehat G_\gamma+
 \bigO\!\left(\sqrt{\frac{D^2R^2d\log(dD)V(W)+\log(Dmn/\delta)}{(mn-1)\gamma^2}}\right).
\end{equation}
Here $\widehat G_\gamma$ is the empirical fraction of examples whose true-class logit fails to exceed the largest competing logit by $\gamma$, using the \emph{same} perturbed global model on every example. An average of margins from differently perturbed local networks measures a different predictor. Curvature measurements in Section~\ref{sec:evidence} probe the geometry empirically. They do not supply the spectral complexity or the margins appearing in \eqref{eq:margin}.

%% file: sections/05-duals.tex
\section{Keeping multipliers current during absence}\label{sec:dual}
A consensus formulation ties each client copy $u_i$ to the global $\w$. With a common penalty $\tau>0$ and a client-specific multiplier $\lam_i^t$, the local augmented objective is
\begin{equation}\label{eq:al}
 L_i^t(u)=f_i(u)+\inner{\lam_i^t}{u-\w^t}+\tfrac{\tau}{2}\norm{u-\w^t}^2.
\end{equation}
At an exact local solution, $\nabla f_i(u_i)+\lam_i^t+\tau(u_i-\w^t)=0$. A dual ascent update $\lam_i^{t+1}=\lam_i^t+\tau(u_i-\w^t)$ then tracks $-\nabla f_i(u_i)$. Dynamic regularization and federated ADMM exploit this local balance \cite{zhang2021fedpd,acar2021federated,gong2022fedadmm,wang2022fedadmm}. Holding $\lam_i$ fixed while the broadcast model changes breaks the temporal link. Upon returning, a client can solve its current subproblem accurately and still be guided by an old multiplier. Increasing $K$ treats local-solver error directly, while its effect on multiplier age is indirect.

\lead{A virtual update from the active cohort} Let $x_i^t$ be the active client's local endpoint and let $\bar x^t=m_p^{-1}\sum_{i\in\cP_t}x_i^t$. \method{A-FedPD} performs the updates \cite{sun2024fedpd}
\begin{equation}\label{eq:dual}
 \lam_i^{t+1}=\lam_i^t+\tau(z_i^t-\w^t),\qquad
 z_i^t=\begin{cases}x_i^t&i\in\cP_t,\\ \bar x^t&i\notin\cP_t,\end{cases}
 \qquad \w^{t+1}=\bar x^t+\bar\lam^{t+1}/\tau ,
\end{equation}
where $\bar\lam=m^{-1}\sum_i\lam_i$. Every inactive client receives the same virtual increment on the server. If its counterfactual endpoint were $x_{i,\rm full}^t$, the imputation error in that increment would be $\tau(\bar x^t-x_{i,\rm full}^t)$. A cohort drawn predominantly from one population can be a poor surrogate for another. The construction repairs the missing shared movement while retaining each client's past dual offset.

\lead{Exact bookkeeping at population scale} Averaging \eqref{eq:dual} gives $\bar\lam^{t+1}=\bar\lam^t+\tau(\bar x^t-\w^t)$ because the active deviations from $\bar x^t$ sum to zero. A server can store one common vector $c^t$ and a residual $r_i^t=\lam_i^t-c^t$ per known client. Increment $c^t$ by $\tau(\bar x^t-\w^t)$ and only change active residuals, by $\tau(x_i^t-\bar x^t)$. Substitution into $c^{t+1}+r_i^{t+1}$ recovers both branches of \eqref{eq:dual}. This implementation reproduces the mathematical iterates, updates $O(m_p d)$ vector entries per round, and uses $O(md)$ storage in the worst case for $d$ model parameters. It requires stable client identities. It also illustrates how a consistency requirement can guide an implementation decision without changing the optimizer.

\begin{theorem}[Inexact primal--dual subproblems, \cite{sun2024fedpd}]\label{thm:dual}
Let $e_i^t=\nabla L_i^t(x_i^t)$ be the local first-order residual. For $L$-smooth local losses, the proof takes $\tau>6L$ sufficiently large for its descent coefficient to be positive. The cited bound uses the virtual-sequence displacement estimate stated in Appendix~\ref{app:dual} and a conditional mean-zero residual with $\E_t\norm{e_i^t}^2\leq\varepsilon_{\rm loc}$. For $\bar\w^{t+1}=\bar x^t$ it obtains
\begin{equation}\label{eq:dual-rate}
 \frac1T\sum_{t=1}^T\E\norm{\nabla f(\bar\w^t)}^2
 \ \leq\ \frac{f(\bar\w^1)-\inf f+c_\tau R_0}{a_\tau T}+\bigO(\varepsilon_{\rm loc}),
\end{equation}
where $a_\tau,c_\tau>0$ are method constants and $R_0$ is the initial mean squared local displacement.
\end{theorem}

Appendix~\ref{app:dual} also derives the same order directly from the implemented updates for $m_p\ge2$ and a sufficient penalty, and gives an exact one-client counterexample. With a fixed local tolerance the bound retains a residual term. A small number of SGD steps alone establishes neither conditional unbiasedness of the subproblem residual nor its size. Near stationarity, local residual estimates and cohort representativeness answer different diagnostic questions.

For a local step of size $\eta$, the previous offset $u-\w^t$ enters the next offset multiplied by $1-\eta\tau$. With $0<\eta\tau<1$ this damps the inherited offset. The local gradient and multiplier still drive the update. With bounded smooth Lipschitz losses $0\leq\ell\leq H$, equal client sizes and a round-wise step $\eta_t=\eta_0/(t+1)$ with admissible $\eta_0=O(1/K)$, under the round-exposure convention in Appendix~\ref{app:dual}, the accompanying stability analysis bounds the signed expected generalization gap by $C(Hm_pT)^{a/(1+a)}/(mn)$, where $a=\mu L$ and $\mu$ is the schedule constant \cite{sun2024fedpd}. Counting all $K$ independent local draws changes the early-exposure term. The stationarity comparison also depends on how well the local subproblem is solved.

Splitting methods make this distinction concrete. FedSplit and randomized Douglas--Rachford methods allocate separate states to local and consensus operators \cite{pathak2020fedsplit,tran2021feddr}. Randomized proximal updates trade frequency of communication against local progress \cite{mishchenko2022proxskip,condat2023randprox}. Client sampling, compression and importance sampling change which residuals the coordinator can reconstruct \cite{grudzien2023can,grudzien2023improving}. Hybrid local solvers and variance-reduced ADMM variants tailor the local minimization or its correction \cite{niu2023fedhybrid,wang2023beyond,sun2023fedspeed}. Composite objectives, convex side constraints and primal--dual fixed-point methods introduce further feasibility or proximal conditions \cite{yuan2021federated,he2024federated,zhu2024federated}. In each family a suitable diagnostic is the first-order or fixed-point residual of its own subproblem.

%% file: sections/06-topology.tex
\section{When the network sets the reference}\label{sec:topology}
In server-coordinated FL (CFL), participating clients receive one shared model at the next broadcast. In decentralized FL (DFL), each node reconstructs its next initialization from its neighbors \cite{lalitha2018fully,sun2022decentralized,lian2017can,koloskova2020unified,wang2021cooperative}. Let $X\in\mathbb R^{m\times d}$ stack the client endpoints as rows, $J=\mathbf1\mathbf1\trans/m$ the global averaging projection, and $A$ a symmetric, nonnegative, doubly stochastic matrix on a connected graph. Require $\lambda=\norm{A-J}_2<1$. One round of mixing preserves the mean $JAX=JX$, but each client trains from its own mixed row. For $q$ exchanges of a fixed set of endpoints,
\begin{equation}\label{eq:gossip}
 \norm{(I-J)A^qX}_F=\norm{(A-J)^q(I-J)X}_F\leq\lambda^q\norm{(I-J)X}_F.
\end{equation}
This follows from $AJ=JA=J$ and submultiplicativity \cite{montenegro2006mathematical}. The conserved mean does not eliminate optimization error. In the next local step, gradients are evaluated at different mixed rows, as in \eqref{eq:path}. More exchanges, better weights, or a shorter local interval can reduce the resulting error by different means. Repeated gossip, gradient tracking, compression and bias correction use different communication resources to address it \cite{hashemi2021benefits,shi2015extra,yuan2020influence,koloskova2019decentralized,alghunaim2022unified}.

\begin{theorem}[Stability under sampling and mixing, \cite{sun2025towards}]\label{thm:network}
Assume $n$ examples per client, bounded $G$-Lipschitz and $L$-smooth sample losses in $[0,U]$, and unbiased stochastic gradients with local variance bounded by $\sigma_l^2$. Let the step at local iteration $s=tK+k$ decay as $\mu_0/(s+s_0)$ with $a=\mu_0L\in(0,1]$ and the offsets chosen as in the source analysis. Use the source coupling in which the conditional expected change of a gradient at a replaced example is bounded by $2\sigma_l$ at a common point (the bounded-gradient alternative is given in Appendix~\ref{app:network}). Write $\theta=a/(1+a)$ and collect the common factor into $C_T=(4/n)(\sigma_l G/L)^{1/(1+a)}(UTK)^\theta$. The balanced-cutoff stability upper bounds for uniform cohort sampling in CFL and full participation on a fixed symmetric graph in DFL are, respectively,
\begin{equation}\label{eq:top-bounds}
 B_{\rm CFL}=C_T\frac{m_p^\theta}{m},\qquad
 B_{\rm DFL}=C_T\left(\frac{1+6\sqrt m\,\kappa_\lambda}{m}\right)^{1/(1+a)}.
\end{equation}
\end{theorem}

For $0<\lambda<1$ and $\nu=1-a$, $\kappa_\lambda$ is an upper bound, uniform in $t\geq1$, on $t^\nu\sum_{s=0}^{t-1}\lambda^{t-1-s}(s+1)^{-\nu}$. It measures how long past local updates survive mixing. Appendix~\ref{app:network} gives the finite-schedule cutoff form. The source sets its network penalty to zero for exact averaging. In that case \eqref{eq:top-bounds} agrees with the bound for full-participation CFL. At fixed $n,T,K$, the displayed decentralized bound tends to zero with increasing $m$ when $\kappa_\lambda=o(\sqrt m)$. A family of graphs needs sufficient mixing as it grows. Equation~\eqref{eq:top-bounds} orders the stability guarantees. Ranking observed test risks also requires data, a training horizon and a bandwidth budget. Alternative bounds for decentralized SGD can even remove a network dependence for a different output or stability notion \cite{sun2021stability,bars2023improved,wang2024towards}.

In CFL the factor $m_p^\theta$ increases at fixed $T,K$, whereas a larger cohort can reduce optimization noise. For illustration, if an empirical optimization term scales as $A_0/m_p$ and a stability term as $B_0m_p^\theta/m$, minimizing their sum gives $m_p=(A_0 m/(\theta B_0))^{1/(1+\theta)}$ before imposing the integer and capacity limits. This is a conditional scaling argument. Its constants and assumed optimization law change with the task. In DFL the corresponding operating choices are the graph, its weights and $q$. To reduce a fixed disagreement by a factor $r\in(0,1)$, \eqref{eq:gossip} gives the sufficient budget $q\geq\lceil\log(r)/\log(\lambda)\rceil$ for $0<\lambda<1$.

Topology can also be designed. Dynamic consensus, segmented or sparsified gossip and empirical gossip comparisons study the information exchanged at a node \cite{chen2022dacfl,hu2019decentralized,tang2022gossipfl,hegedHus2021decentralized}. Balancing data across neighborhoods can reduce disagreement before mixing \cite{bellet2022d,shi2023improving}. Heterogeneity-aware overlays, graph embedding and energy-aware topology optimization trade physical links against mixing and availability \cite{cao2021hadfl,hua2022efficient,duan2024topology,liu2025towards}. Hierarchical resource allocation, trusted communication and network-structured learning impose additional constraints on feasible links \cite{lim2021decentralized,gholami2022trusted,sarcheshmehpour2021federated}. A dual constraint can directly control disagreement in decentralized ADMM \cite{li2025dfedadmm}. Conversely, some disagreement acts as a form of regularization in analyses relating decentralized SGD to average-direction SAM \cite{zhu2023decentralized}. An agreement metric acquires its meaning from the intended target.

%% file: sections/06a-synthesis.tex
\par\ifdim\dimexpr\pagegoal-\pagetotal\relax<.46\textheight\newpage\fi
\section{What consistency changes, and where it helps}\label{sec:synthesis}
Consistency enters the five analyses at different points in a training round. For the first two, the central question is whether the average message gives a useful descent direction for $f$. For sharpness, independently maximizing before averaging changes the objective itself. For primal--dual training and gossip, the reference is a constraint that must continue to hold as clients or their neighbors move. Table~\ref{tab:scope} summarizes the resulting guarantees.

\begin{table}[H]
\caption{What the correction controls and the scope of the theoretical comparison. The theorem numbers refer to the restatements in this article.}\label{tab:scope}
\centering\small\setlength{\tabcolsep}{4pt}\renewcommand{\arraystretch}{1.16}
\begin{tabularx}{\linewidth}{@{}>{\raggedright\arraybackslash}p{.16\linewidth}Y Y@{}}\toprule
Mechanism & Quantity that can improve & Scope of the result \\\midrule
Relaxed start & Reversing a historical offset can reduce endpoint drift. This changes the divergence contribution in a stability analysis. & Theorem~\ref{thm:init} retains the local-SGD order under its history estimate. The sharper stability comparison depends on the schedule and horizon.\\
Adaptive direction & A shared direction scales the within-step dispersion by $\alpha^2$. A common denominator preserves cancellation at a stationary point. & Theorem~\ref{thm:lada} measures the auxiliary iterate. Local moments and a stale shared direction can both influence the aggregate.\\
Sharpness & Aligning perturbations reduces the discrepancy between local and shared worst-case objectives. & Theorem~\ref{thm:smoo} uses exact virtual solves and a shrinking radius. The margin bound applies to one global predictor.\\
Dual history & A virtual increment follows population movement during absence. It preserves a client's past dual offset. & Theorem~\ref{thm:dual} retains a local-solver error floor. The active mean only approximates a missing client's displacement.\\
Peer mixing & Each exchange contracts existing disagreement by at most $\lambda$. This limits its propagation across rounds. & Theorem~\ref{thm:network} compares stability bounds for a specified output and graph. Extra exchanges incur a link cost.\\\bottomrule
\end{tabularx}
\end{table}

In practice, a correction is worth considering when the discrepancy it addresses is visible at the operation that uses the state. Two clients can disagree at initialization and agree more closely at upload. They can also agree on a model and still produce a biased normalized direction. The distinction suggests holding the task and the resource budget fixed while changing one operation and measuring both its discrepancy and the learning outcome. The following experiments illustrate when those measurements agree and when they part company.

%% file: sections/07-evidence.tex
\section{Experimental checks of consistency}\label{sec:evidence}
A controlled example lets us observe the global reference directly. We then revisit published vision experiments, where the reference is usually available only through a proxy. Each comparison uses the dataset, architecture and evaluation rule of its own study.

\subsection{A controlled test of direction bias}
Give each of $m=12$ equally weighted clients a quadratic loss $f_i(w)=(w-a_i)^\top H_i(w-a_i)/2$ in $d=20$ dimensions. Draw the centers independently from $\mathcal N(0,I)$ and the diagonal curvatures independently as $\exp(U[-\log 4,\log 4])$. The exact global minimizer is $w^\star=(\sum_i H_i)^{-1}\sum_i H_i a_i$. Write $P_i=H_i^{-1}$ and $\bar P=m^{-1}\sum_iP_i$. Vary only how much of the local scaling is replaced by the common scaling,
\begin{equation}\label{eq:control-dir}
 u_\gamma(w)=\frac1m\sum_i\big[(1-\gamma)P_i+\gamma\bar P\big]\nabla f_i(w),\qquad 0\leq\gamma\leq1 .
\end{equation}
All clients take one exact step per round, with $w^0=0$, $\eta=0.25$ and $T=100$. Every value of $\gamma$ uses the same 20 problem instances (seeds 0--19). Equation~\eqref{eq:pcov} predicts that the residual relative to $\bar P\nabla f(w)$ is multiplied by $1-\gamma$ at any fixed $w$. We measure it at $w^\star$ and separately measure the objective reached after training.

\begin{table}[H]
\caption{Changing only the diagonal scaling in \eqref{eq:control-dir}. Entries are the mean and sample standard deviation across 20 paired problem instances. The last column is $100[f(w^T)-f(w^\star)]/[f(w^0)-f(w^\star)]$. The final row reports an upper threshold for numerical residuals.}\label{tab:scaling-check}
\centering\small\renewcommand{\arraystretch}{1.10}
\input{experiments/scaling_table}
\end{table}

\begin{figure}[htb]
\centering\includegraphics[width=.78\linewidth]{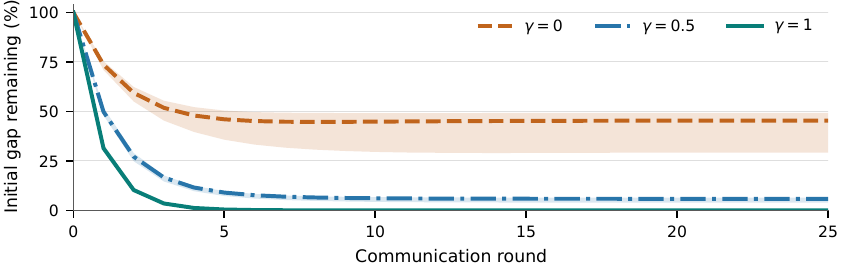}
\caption{Objective gap during the first 25 rounds of the quadratic experiment. Curves show means and ribbons give the 25th--75th percentiles over the same 20 problems at each round. Local scaling alone can descend quickly and then settle above the minimum of the global objective. Table~\ref{tab:scaling-check} gives the endpoints at round 100.}\label{fig:scaling-paths}
\end{figure}

With entirely local scaling, the aggregate update points toward the unweighted mean of the centers and leaves 45.36\% of the initial objective gap on average. At $\gamma=0.50$ the reference residual halves and the remaining gap is 5.85\% (Table~\ref{tab:scaling-check}). As a control, substituting the same $H_i=\bar H$ at every client, with the centers and mean curvature held fixed, gives both measures below $10^{-12}$ for every $\gamma$. Figure~\ref{fig:scaling-paths} shows that the gap emerges early and persists after local progress slows. Lemma~\ref{lem:synthetic} gives the exact transient and limiting point for each curve. The next comparison examines learned adaptive moments and test accuracy on vision data.

\subsection{Evidence from vision benchmarks}
The studies use different backbones, participation schemes, stopping rules and local optimizers \cite{sunthesis}. Differences below are percentage points (pp) unless otherwise indicated. Reported $\pm$ values retain the source's dispersion notation.
\begin{table}[H]
\caption{Within-study test accuracy (\%). Each row compares the stated methods under the same dataset and partition. Dirichlet concentration is denoted by $u$. The last column gives the difference between reported means.}\label{tab:accuracy}
\centering\small\renewcommand{\arraystretch}{1.13}\setlength{\tabcolsep}{4.3pt}
\begin{tabularx}{\linewidth}{@{}Y l r r r@{}}\toprule
Comparison & Dataset / split & Baseline & Corrected & $\Delta$ (pp)\\\midrule
\method{FedAvg} $\to$ \method{FedInit} & CIFAR-10 / $u=0.1$ & $72.53\pm.17$ & $75.95\pm.19$ & $+3.42$\\
\method{FedAvg} $\to$ \method{FedInit} & CIFAR-100 / $u=0.1$ & $42.62\pm.22$ & $44.22\pm.21$ & $+1.60$\\\addlinespace[3pt]
\method{MoFedSAM} $\to$ \method{FedSMOO} & CIFAR-100 / $u=0.6$ & $54.38\pm.22$ & $53.92\pm.18$ & $-0.46$\\
\method{MoFedSAM} $\to$ \method{FedSMOO} & CIFAR-100 / $u=0.1$ & $44.85\pm.25$ & $46.48\pm.13$ & $+1.63$\\\addlinespace[3pt]
\method{FedDyn} $\to$ \method{A-FedPD} & CIFAR-10 / $u=0.1$ & $79.44\pm.25$ & $81.05\pm.28$ & $+1.61$\\
\method{FedDyn} $\to$ \method{A-FedPD}$^\dagger$ & CIFAR-100 / $u=0.1$ & $46.30\pm.26$ & $48.15\pm.28$ & $+1.85$\\\bottomrule
\end{tabularx}
\par\vspace{3pt}\raggedright\footnotesize All rows use 10 of 100 clients. \method{FedInit} uses ResNet-18-GN and three runs. \method{FedSMOO} uses ResNet-18-GN and five runs. \method{A-FedPD} uses four seeds, ResNet-18 for CIFAR-10 and LeNet ($\dagger$) for CIFAR-100. Sources \cite{sun2023understanding,shen2026rethinking,sun2023dynamic,sun2024fedpd}.
\end{table}

\lead{Historical endpoints} The \method{FedInit} experiments use Dirichlet label partitions \cite{hsu2019measuring}, including sampling with replacement. For CIFAR-10 and CIFAR-100 \cite{krizhevsky2009learning} they run 500 and 800 rounds, respectively, smooth the accuracy curve and take the largest value in the last 50 rounds. Table~\ref{tab:accuracy} gives the ResNet-18-GN results under strong heterogeneity. In the CIFAR-100 setting SCAFFOLD reaches $44.41\%$, slightly above \method{FedInit}'s $44.22\%$. Table~\ref{tab:relaxation} directly compares the terminal divergence and accuracy as the initialization coefficient changes. At $\beta=0.10$ the two measures improve together relative to $\beta=0$. Moving the coefficient to $-0.10$ also lowers divergence, from $0.855$ to $0.815$, yet accuracy falls by $5.06$ pp. A smaller terminal distance can accompany either outcome. On CIFAR-10, averaged over the reported V100 round timings with 10 of 100 active clients, the extra initialization changes $19.38$ to $20.41$ seconds per round. With 10 of 200 active clients the corresponding times are $15.87$ and $16.14$ seconds. The wire format is the same in each pair.

\begin{table}[htb]
\caption{The complete CIFAR-10 initialization sweep at 500 rounds, with 10 of 100 clients and Dirichlet concentration 0.1. Columns vary only $\beta$. The source reports $\infty$ for the final divergence at $\beta=0.15$ \cite[Table~3]{sun2023understanding}.}\label{tab:relaxation}
\centering\small\renewcommand{\arraystretch}{1.08}\setlength{\tabcolsep}{5.2pt}
\begin{tabular}{@{}lrrrrrrrr@{}}\toprule
$\beta$ & $-0.20$ & $-0.10$ & $0$ & $0.01$ & $0.02$ & $0.05$ & $0.10$ & $0.15$ \\\midrule
Test accuracy (\%) & 64.70 & 67.47 & 72.53 & 72.82 & 73.45 & 74.65 & 75.95 & 44.47 \\
Terminal $\Delta_T$ & 0.873 & 0.815 & 0.855 & 0.875 & 0.850 & 0.823 & 0.760 & $\infty$ \\\bottomrule
\end{tabular}
\end{table}

\begin{table}[tb]
\caption{Communication rounds to reach a specified test accuracy in the \method{FedLADA} study \cite{sun2023efficient}. NR denotes a target unreached during that study's training horizon. Ratios divide the indicated baseline's round count by \method{FedLADA}'s count.}\label{tab:rounds}
\centering\small\renewcommand{\arraystretch}{1.09}\setlength{\tabcolsep}{7pt}
\begin{tabular}{@{}lrrr@{}}\toprule
 & CIFAR-10 & CIFAR-100 & Tiny-ImageNet\\
Test target (\%) & 84 & 51 & 38\\\midrule
\method{SCAFFOLD} & 1,177 & 677 & 494\\
\method{FedCM} & 1,211 & 721 & 455\\
\method{LocalAdam} & NR & NR & NR\\
\method{FedLADA} & 598 & 447 & 269\\\midrule
Fastest listed successful baseline / \method{FedLADA} & 1.97 & 1.51 & 1.69\\\bottomrule
\end{tabular}
\end{table}

The wider initialization study also tests VGG-11 \cite{simonyan2014very}. With heterogeneous feature statistics, local normalization choices can themselves affect the result \cite{gn_bn,li2021fedbn}. Keeping the backbone and normalization fixed in a paired comparison prevents such a change from being attributed to initialization.

\lead{Local adaptive training} The \method{FedLADA} study uses ResNet-18, 10\% participation, batch size 50 and five local epochs on CIFAR-10 or two on the other datasets. The unamended local adaptive optimizer reaches 98\% CIFAR-10 training accuracy in 575 rounds. \method{FedLADA} takes 765. Only the latter reaches the 84\% test target among this pair (Table~\ref{tab:rounds}). On Tiny-ImageNet, at the reported 1,500-round comparison, test accuracy varies from $29.5\%$ at $\alpha=0.01$ to $41.9\%$ at $0.05$ and $34.1\%$ at $1$. This reversal at very small $\alpha$ is consistent with an overly stale $q^t$ in \eqref{eq:lada}. The sweep measures the effect of the mixing coefficient, without directly measuring the direction error. Round counts also omit the extra optimizer state per transmission.

\begin{figure}[htb]
\centering\includegraphics[width=\linewidth]{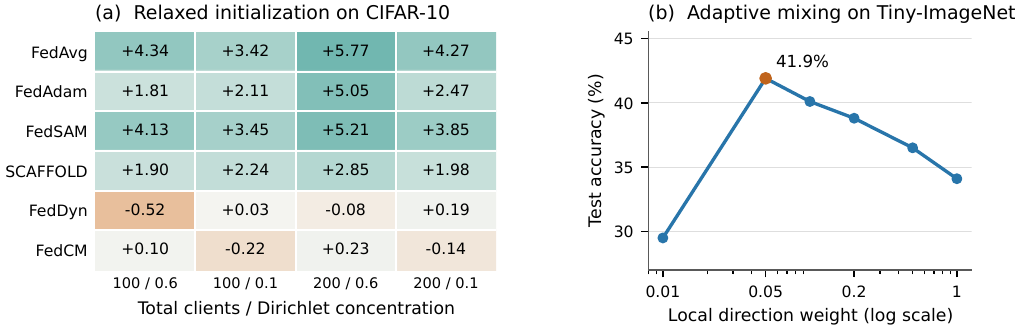}
\caption{Two within-study interventions. Left, each cell is the change in test accuracy (percentage points) after adding relaxed initialization to the named optimizer on CIFAR-10/ResNet-18-GN. All columns use ten active clients. Right, points are the reported Tiny-ImageNet accuracies after 1,500 rounds for six values of the local-direction weight. The line connects the evaluated settings \cite{sun2023understanding,shen2026rethinking,sun2023efficient,sunthesis}.}\label{fig:paired-interventions}
\end{figure}

Figure~\ref{fig:paired-interventions} adds two controls. Holding the baseline fixed, relaxed initialization adds $3.42$--$5.77$ pp to \method{FedAvg} across the four settings but changes \method{FedDyn} by $-0.52$ to $+0.19$ pp. Holding the adaptive method fixed, both extremes of the direction weight fall below its best observed interior value. These comparisons identify dependence on the baseline and on the weight. Neither study directly measures the covariance term in \eqref{eq:pcov}.

\lead{A shared perturbation} In the CIFAR-10 three-class-per-client ablation, adding dynamic regularization to the baseline changes $74.08$ to $77.63\%$. Adding SAM gives $80.13\%$, and adding perturbation correction gives $81.58\%$ \cite[Table~8]{sun2023dynamic}. The final incremental gain of 1.45 pp isolates the correction more closely than a comparison of the complete algorithms. In a separate CIFAR-10 Dirichlet-$0.1$ experiment, \method{FedSAM} and \method{FedSMOO} report endpoint divergence of 1.245 and 1.061, with leading Hessian eigenvalues of 177.18 and 107.44 \cite{sun2023dynamic,sunthesis}. The measured divergence is model-space distance to the aggregate. It is a different observable from the perturbation gap in \eqref{eq:sharp-gap}. On CIFAR-100 the comparison with \method{MoFedSAM} changes sign with heterogeneity (Table~\ref{tab:accuracy}). Those experiments use 800 rounds and do not use the \method{FedInit} best-of-window reporting rule.

\lead{Inactive client state} The \method{A-FedPD} comparison fixes $K=50$, $T=800$, and 10\% participation \cite{sun2024fedpd}. The four-seed accuracy results in Table~\ref{tab:accuracy} include one comparison for each backbone. On ResNet-18/CIFAR-100 with Dirichlet-$0.1$, \method{FedDyn} is marked unstable under the reported configuration, while \method{A-FedPD} reaches $46.24\pm0.26\%$. Another baseline, SCAFFOLD, reaches $47.12\pm0.34\%$ in the same setting. The comparison is consistent with a benefit from updating inactive multipliers, although the baseline also differs in its local solver. A direct attribution would require freezing and updating the same multipliers within one solver.

\begin{figure}[tb]
\centering\includegraphics[width=\linewidth]{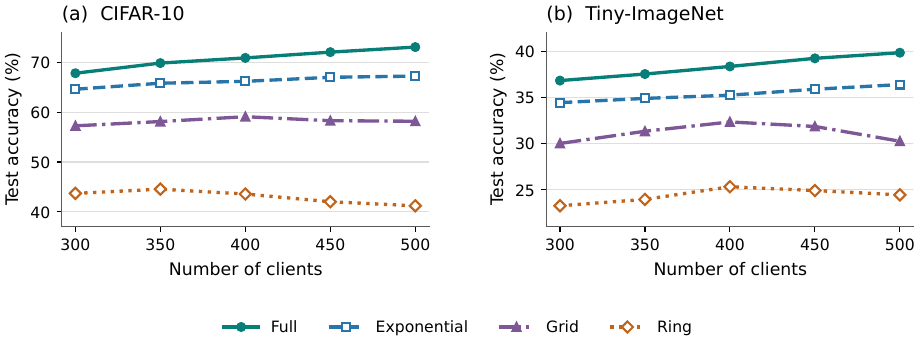}
\caption{Test accuracy as the decentralized population grows, with 100 training examples per client. Every marker is a tabulated value from the topology study \cite{sun2025towards,sunthesis}. Lines connect the observations. Vertical scales differ between panels.}\label{fig:topology}
\end{figure}

\lead{Scaling the graph} In Figure~\ref{fig:topology} each additional client contributes training data. From 300 to 500 clients, fully connected mixing improves CIFAR-10 accuracy from $67.86$ to $73.12\%$. Ring mixing changes it from $43.68$ to $41.19\%$. On Tiny-ImageNet the ring peaks at $25.32\%$ at 400 clients and ends at $24.44\%$, still above its $23.26\%$ starting point. Thus weak mixing can erase part of the benefit of additional data without producing a monotone decline. A separate fixed-dataset, three-seed comparison illustrates the role of the bandwidth convention. At $m=100$, $K=50$ on CIFAR-10, a grid scores $73.27\pm0.19\%$, centralized averaging with five clients scores $73.13\pm0.37\%$, and doubling server capacity to ten active clients gives $75.48\pm0.44\%$ \cite{sun2025towards}. These are capacity-matched operating points as defined in the study. An end-to-end deployment would also need measured link rates and straggler times.

%% file: experiments/scaling_table.tex
\begin{tabular}{@{}c c c@{}}\toprule
Shared weight $\gamma$ & Residual $\norm{u_\gamma(w^\star)}/\sqrt d$ & Initial objective gap remaining (\%) \\\midrule
0.00 & $0.211\,\pm\,0.033$ & $45.36\,\pm\,25.51$ \\
0.25 & $0.158\,\pm\,0.024$ & $17.73\,\pm\,9.69$ \\
0.50 & $0.106\,\pm\,0.016$ & $5.85\,\pm\,3.13$ \\
0.75 & $0.053\,\pm\,0.008$ & $1.13\,\pm\,0.60$ \\
1.00 & $<10^{-12}$ & $<10^{-12}$ \\
\bottomrule
\end{tabular}

%% file: sections/08-deployment.tex
\section{Implications for a training system}\label{sec:deploy}
A federated training design must decide where each reference is maintained, how old it may be when used, and what it costs to refresh it. The same questions arise for several existing optimizers. Table~\ref{tab:system} accounts for the state exchanged in the direct implementations studied here. The baseline is synchronous \method{FedAvg}. Each full vector has $d$ entries. Counts are per active client per round, exclude protocol headers and compression, and treat a broadcast as a logical download to each receiver.

\begin{table}[H]
\caption{State placement and communication in a direct implementation. Down/up counts are model-sized vectors. Persistent state excludes the base model and transient gradient buffers.}\label{tab:system}
\centering\small\renewcommand{\arraystretch}{1.16}\setlength{\tabcolsep}{4pt}
\begin{tabularx}{\linewidth}{@{}l c Y Y@{}}\toprule
Method & Down/up & Persistent state or reference & Main cost or constraint \\\midrule
\method{FedAvg} & 1 / 1 & Server model and active endpoint mean & $K$ local gradient evaluations.\\
\method{FedInit} & 1 / 1 & Last local endpoint at each returning client & One cached vector per client, plus an extrapolation at entry.\\
\method{FedLADA} & 3 / 2 & Server model, second-moment estimate and previous direction & Reset first moment, average client second moments, broadcast the amendment.\\
\method{FedSMOO} & 2 / 2 & Shared perturbation and mean model multiplier on server, two multiplier vectors at client & Two gradient evaluations per local step and a perturbation message.\\
\method{A-FedPD} & 2 / 1 & Current client multiplier on server, optionally stored as in Section~\ref{sec:dual} & $O(md)$ worst-case server state. Virtual updates run on the server.\\
Decentralized averaging & Peer links & Each node's model and the graph weights & $q\deg(i)$ point-to-point sends per node for $q$ mixing steps, excluding self-links.\\\bottomrule
\end{tabularx}
\end{table}

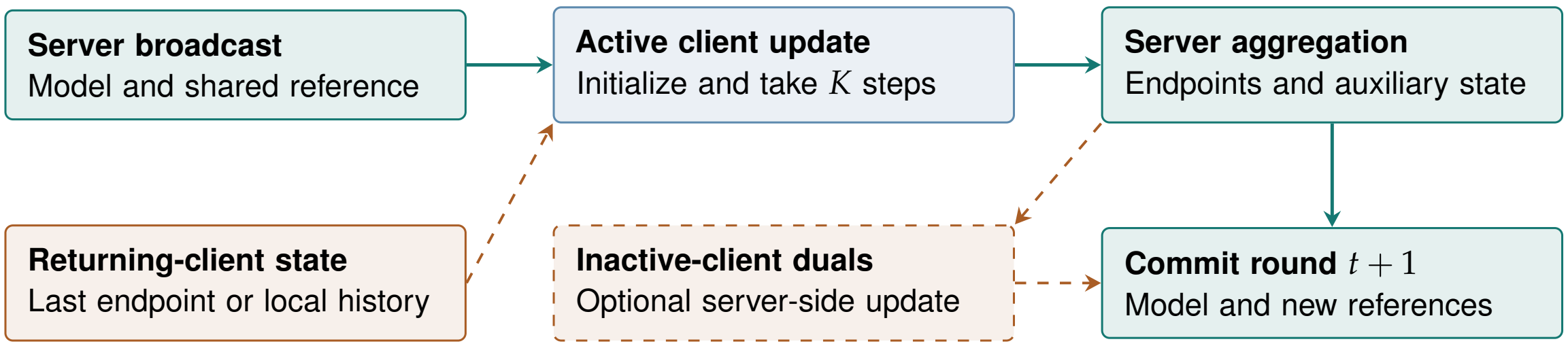
\begin{figure}[H]
\centering\resizebox{.97\linewidth}{!}{\input{figures/round}}
\caption{Where a correction enters a server-coordinated round. Solid arrows are the common model-update path. Dashed arrows show optional dependencies. Historical initialization reads returning-client state. Virtual dual correction uses the active aggregate to update server records for absent clients before the round commits.}\label{fig:round}
\end{figure}

\lead{State used in one training round} Figure~\ref{fig:round} gives the state dependencies. The server freezes the model, cohort and any shared reference for the round. A selected client loads only the state associated with that model version, constructs its initial condition and performs the local solve. It returns the endpoint and the auxiliary message specified by the selected method. The server aggregates valid replies, updates the reference, and commits their common version for the next round. The update to inactive multipliers in \eqref{eq:dual} takes place at this commit. If a selected client drops out, the effective cohort and any sample weights must correspond to the messages actually used. A decentralized round replaces the server commit with the specified neighbor mix. Its observable is the post-mixing disagreement in \eqref{eq:gossip}. The method-specific guarantees above assume their respective synchronous sampling rules.

\lead{Choosing under a budget} For $b$ bytes per coordinate and $D_\downarrow,D_\uparrow$ vectors, a direct unicast implementation moves $m_pbd(D_\downarrow+D_\uparrow)$ bytes per round across the server links. With $d=10^6$, 32-bit values and ten active clients, \method{FedInit} moves 80 MB and \method{FedLADA} 200 MB per round (decimal MB). The same accounting can be applied to Table~\ref{tab:rounds}. Direct \method{SCAFFOLD} sends a model and a control variate down and returns their two increments \cite{karimireddy2020scaffold}. To reach 84\% on CIFAR-10, the resulting payload ratio is $598\times5/(1177\times4)=0.635$ for \method{FedLADA} relative to \method{SCAFFOLD}. With uncompressed unicast at equal precision this means 36.5\% fewer bytes to that target. A common downlink broadcast, cached side information or asymmetric link prices changes this accounting. Let $T_\star$ denote rounds to a prespecified validation target. If the uplink is the bottleneck, compare $T_\star m_pbdD_\uparrow$. If stragglers dominate, compare the sum of the slowest active-client time plus server and communication time in each round. A stationarity rate or a reduction in $T_\star$ alone cannot choose between those regimes. Efficient communication, sketching and synchronization studies motivate measuring these quantities separately \cite{asad2020fedopt,ivkin2019communication,wang2022addressing}.

\lead{Selecting a correction} If clients return frequently and the uplink is fixed at one model vector, historical endpoints allow a direct trial of \eqref{eq:ri}. Log the endpoint distance when a client returns, its subsequent upload distance and the validation metric at a fixed number of transmitted bytes. If adaptive clients fit training data rapidly while the global validation metric stalls, compare several $\alpha$ values against both a local-adaptive and a server-adaptive baseline, with their respective payloads. If the deployed objective is robustness to a common perturbation, test the difference between shared and separate margins before paying for the second gradient. When a primal--dual solver deteriorates after absences, stratify its first-order residual by the number of missed rounds and compare virtual with frozen multipliers at the same local-solver tolerance. For a fixed peer graph, measure the attainable disagreement reduction per exchange before increasing $q$ or admitting more nodes.

These choices are alternative or, in specific cases, compatible interventions. Implementing all four corrections simultaneously would change the effective objective, timing and state. The reported \method{A-FedPD} experiments do include a SAM variant \cite{sun2024fedpd}. Other combinations require their own evaluation. For a new composition a useful first experiment retains the baseline, adds each mechanism alone, and then measures their joint effect with identical partitions, seeds, target accuracy and resource budget. The resulting comparison would test whether a measured discrepancy helps choose a correction under a fixed budget.

%% file: figures/round.tex
\definecolor{roundserver}{HTML}{157872}
\definecolor{roundclient}{HTML}{356D9A}
\definecolor{roundstate}{HTML}{A95D25}
\begin{tikzpicture}[
  box/.style={draw=roundclient!80,rounded corners=2pt,line width=.55pt,align=left,text width=3.9cm,inner xsep=6pt,inner ysep=6pt,font=\sffamily\small,text=black},
  flow/.style={-{Stealth[length=4.3pt,width=4.6pt]},draw=roundserver,line width=.9pt},
  optional/.style={-{Stealth[length=4.3pt,width=4.6pt]},draw=roundstate,dashed,line width=.7pt}]
  \node[box,fill=roundserver!11,draw=roundserver] (broadcast) at (0,2.05) {\textbf{Server broadcast}\\Model and shared reference};
  \node[box,fill=roundstate!9,draw=roundstate] (cache) at (0,0) {\textbf{Returning-client state}\\Last endpoint or local history};
  \node[box,fill=roundclient!10] (solve) at (5.15,2.05) {\textbf{Active client update}\\Initialize and take $K$ steps};
  \node[box,fill=roundserver!11,draw=roundserver] (aggregate) at (10.3,2.05) {\textbf{Server aggregation}\\Endpoints and auxiliary state};
  \node[box,dashed,fill=roundstate!9,draw=roundstate] (inactive) at (5.15,0) {\textbf{Inactive-client duals}\\Optional server-side update};
  \node[box,fill=roundserver!11,draw=roundserver] (commit) at (10.3,0) {\textbf{Commit round $t+1$}\\Model and new references};
  \draw[flow] (broadcast.east) -- (solve.west);
  \draw[flow] (solve.east) -- (aggregate.west);
  \draw[optional] (cache.east) -- (solve.south west);
  \draw[flow] (aggregate.south) -- (commit.north);
  \draw[optional] (aggregate.south west) -- (inactive.north east);
  \draw[optional] (inactive.east) -- (commit.west);
\end{tikzpicture}

%% file: sections/09-discussion.tex
\section{Scope and open questions}\label{sec:scope}
\lead[?]{Which quantities should remain local} For statistical heterogeneity, a global target can specify agreement in only part of the model \cite{ye2023heterogeneous}. Local batch normalization, restricted softmax, logits calibration and neuron-wise rates each put client specificity in a different component \cite{li2021fedbn,li2021fedrs,zhang2022federated2,wang2024fednlr}. Adaptive personalization and knowledge fusion make this choice explicit in the predictor \cite{deng2020adaptive,huang2023fusion}. Contrastive features, shared prototypes and semantic anchors use representation spaces in which agreement can remain meaningful across different client models \cite{li2021model,tan2022fedproto,zhou2025fedsa}. Distillation instead compares predictions on a transfer distribution \cite{zhu2021data,lin2020ensemble,zhang2022fine}. If the transfer distribution omits a client group, matching these predictions has limited evidential value for that group. The appropriate residual depends jointly on the representation and the population on which it is evaluated.

For partially personalized networks only the shared parameter block is subject to a global consensus constraint. Decentralized partial-model training and its convergence theory make this separation explicit \cite{shi2023towards,11359328}. A penalty on the full parameter vector would also regularize the components intended to specialize.

\lead[?]{Generalization to whom} Equation~\eqref{eq:excess} evaluates fresh samples from the training mixture. Unseen-client and domain generalization ask about different distributions. Augmenting inputs or representations can expose nuisance variation \cite{yoon2021fedmix,chen2023fraug,liu2021feddg,nguyen2022fedsr}. Cross-client style transfer and attention or style perturbation provide more targeted interventions \cite{chen2023federated,park2023stablefdg}. Gradient agreement across source domains and globally consistent augmentation address yet other discrepancies \cite{wei2024multi,pourpanah2025federated,liu2024fedgca}. A held-out domain can have a small training-mixture residual and a large error. For personalized sharpness or topology-aware transfer, the reference must name the recipient population and its intended predictor \cite{xing2025flexible,ma2024beyond,jamali2022federated}.

\lead[?]{What can be observed} The full-population gradient, a nonparticipant's endpoint and a test-time worst perturbation are usually unavailable. Cohort proxies can be compared on disjoint random cohorts or checked when a client next participates. Such checks measure repeatability and biased availability can still distort them. Directed or time-varying graphs change the averaging measure, as push-sum and directed-graph stability analyses illustrate \cite{assran2019stochastic,liang2026stability}. Federated graph learning also uses topology inside the task, for representation alignment, aggregation or distillation \cite{hu2026fedtopo,li2023fedgta,zhu2024fedtad}. That graph and the physical communication graph can be quite different. Likewise, when the exchanged object is a prompt, an unlearning update or a privacy-protected aggregate, the reference must incorporate the task or privacy mechanism before its residual can be interpreted \cite{lin2023efficient,li2023visual,li2023subspace,sun2026convergent}.

A minimax objective adds another boundary. Stability of distributed or decentralized gradient descent--ascent couples changes in both players \cite{zhu2023stability,zhu2026stability}. The ball-maximization calculation in \eqref{eq:sharp-gap} depends on a bounded perturbation set and a fixed model. Extending it to evolving adversaries requires specifying both players' update order and reference states.

\lead{A testable research question} At a fixed byte and gradient-evaluation budget, does choosing the correction by a premeasured discrepancy predict validation improvement better than choosing by the size of model drift alone? A direct test would hold the partition and returning-client pattern fixed, measure model, direction and dual discrepancies on the same audit rounds, and vary one correction at a time. Correlated reductions in multiple discrepancies would call for factorial interventions. The existing studies report vision workloads and finite synchronized simulations, with method-specific schedules and hardware. They motivate this test and provide its controls. They leave open joint convergence of a composed optimizer, tight lower bounds on topology-dependent generalization, and performance on a live, intermittently connected fleet.

\section{Conclusion}\label{sec:conclusion}
The consistency principle gives a common reading of several familiar federated operations. A local normalization can change the stationary point of an averaged update. A client-specific perturbation can change the objective being optimized. A stored multiplier or a neighborhood average can change the initial condition for the next round. Each example calls for a different reference. The studies reviewed here suggest recording when the discrepancy is measured, which learning quantity it can affect and what it costs to correct. These questions also apply to methods beyond the ones in the thesis. Whether the resulting choice improves a particular deployment remains an empirical question.

%% file: sections/appendix.tex
\appendix
\numberwithin{equation}{section}
\titleformat{\section}{\sffamily\bfseries\Large}{Appendix~\thesection}{.62em}{}
\section{Proofs and scope of the consistency results}\label{sec:derivations}
The notation follows the main text. Throughout the proofs $s=m_p$ is the number of active clients and $m$ is the population size. Except where the weights $p_i$ are displayed explicitly, the client mean is uniform. The virtual endpoint of a client is the endpoint it would have produced if selected, with local randomness sampled independently of the cohort. This convention permits averaging a fixed collection of potential updates over the cohort. It does not require messages from absent clients. We state the source-specific estimates explicitly when using them as lemmas. The elementary arguments and the reductions from these estimates to the five rates are given below.
\input{appendix/a1-foundations}
\input{appendix/a2-initialization}
\input{appendix/a3-adaptivity}
\input{appendix/a4-sharpness}
\input{appendix/a5-duality}
\input{appendix/a6-topology}

%% file: appendix/a1-foundations.tex
\subsection{Common descent and stability arguments}\label{app:common}
Fix the history before drawing the cohort. In this subsection all expectations with subscript $t$ condition on that history. For a vector-valued implemented operation $U$ and a history-measurable reference $r$, put $u=\E_tU$, $b=u-r$ and $v=\E_t\norm{U-u}^2$. Expanding around $u$ gives
\begin{equation*}\label{ap:orthog}
 \E_t\norm{U-r}^2=\norm b^2+v .
\end{equation*}
The cross term vanishes because $\E_t(U-u)=0$. This observation separates a change of objective or direction from the dispersion of an unbiased cohort. A biased reference estimate would contribute to the first term too.

\begin{lemma}[Sampling a finite client population]\label{lem:cohort}
Fix vectors $d_1,\ldots,d_m$, their average $\bar d$, and a uniform subset $S$ of size $s$. For $m>1$ write $\chi_s=(m-s)/(s(m-1))$. Then
\begin{equation}\label{ap:cohort-var}
 \E_S\bar d_S=\bar d,\qquad
 \E_S\norm{\bar d_S-\bar d}^2
   =\chi_s\,\frac1m\sum_i\norm{d_i-\bar d}^2,
 \quad \bar d_S=\frac1s\sum_{i\in S}d_i .
\end{equation}
\end{lemma}
\begin{proof}
Write $I_i=\mathbf1\{i\in S\}$ and $v_i=d_i-\bar d$. Counting singletons and ordered pairs gives $\E I_i=s/m$ and $\E I_iI_j=s(s-1)/(m(m-1))$ for $i\ne j$. In the expansion of $\E\norm{\sum I_iv_i}^2$, the off-diagonal inner products sum to $-\sum_i\norm{v_i}^2$ since $\sum_i v_i=0$. Substituting the two inclusion probabilities and dividing by $s^2$ proves the variance identity. Linearity proves the mean identity.
\end{proof}
When $d_i$ is itself random, first condition on all potential vectors, then average over them. Variation of their population mean remains after the sampling variance is removed. In particular, observing all clients removes the factor $\chi_s$ but leaves any local gradient noise and any systematic direction error.

\begin{lemma}[Descent for a mismatched mean]\label{lem:descent}
Let $f$ be $L$-smooth, $g=\nabla f(\w)$ and $\w^+=\w-hU$. With $r=g$ and finite second moments,
\begin{equation}\label{ap:descent-exact}
 \E_t f(\w^+)\le f(\w)-\tfrac h2\norm g^2-\tfrac h2(1-Lh)\norm u^2
                +\tfrac h2\norm b^2+\tfrac{Lh^2}{2}v .
\end{equation}
\end{lemma}
\begin{proof}
The smooth upper quadratic gives $\E_t[f(\w^+)-f(\w)]\le-h\inner gu+(Lh^2/2)(\norm u^2+v)$. Expanding $\norm{g-u}^2$ gives the identity $2\inner gu=\norm g^2+\norm u^2-\norm b^2$. Collect the terms in $\norm u^2$.
\end{proof}
\lead{Proof of Proposition~\ref{prop:descent}} For $h\le1/L$ the second term in $\norm u^2$ can be discarded. Sum the remaining inequality over $t$, take total expectations and use $f(\w^T)\ge f_{\rm lo}$. Division by $hT/2$ gives \eqref{eq:descent}. With variable $h_t\le1/L$ the same proof yields
\begin{equation*}\label{ap:variable-descent}
 \frac{\sum_t h_t\E\norm{g_t}^2}{\sum_t h_t}
 \le \frac{2(f(\w^0)-f_{\rm lo})+
 \sum_t h_t\E\norm{b_t}^2 + L\sum_t h_t^2\E v_t}{\sum_t h_t}.
\end{equation*}
A small constant-variance term can be downweighted by reducing the step. A fixed directional error enters with the same weight as the quantity on the left. This is the distinction used in the controlled experiment of Section~\ref{sec:evidence}.

\begin{lemma}[Replace-one comparison for a client mixture]\label{lem:replace}
Let $S^{ij}$ replace example $z_{ij}$ by an independent draw from the same $\mathcal D_i$, and couple the algorithmic randomness in $A(S)$ and $A(S^{ij})$. If the expected absolute change in the loss at the replaced evaluation point is bounded by $e_{ij}$, then
\begin{equation}\label{ap:generalization}
 \left|\E[F(A(S))-f(A(S))]\right|\leq
 \sum_i\frac{p_i}{n_i}\sum_{j=1}^{n_i} e_{ij}.
\end{equation}
For $G$-Lipschitz losses in $[0,U]$, a valid bound on each summand is $U\Pr(B)+G\E[\norm{A(S)-A(S^{ij})}\,\mathbf1_{B^c}]$ for any early-hit event $B$.
\end{lemma}
\begin{proof}
Introduce the independent draw $z'_{ij}$. By exchangeability of the old and new examples within client $i$, $\E\ell(A(S);z'_{ij})=\E\ell(A(S^{ij});z_{ij})$. Subtract $\E\ell(A(S);z_{ij})$, average with weights $p_i/n_i$, and use the triangle inequality. On $B$ the absolute loss difference is bounded by $U$. On its complement it is bounded by $G$ times the output distance.
\end{proof}
Only within-client exchangeability is needed here. The result bounds the signed expected generalization gap in \eqref{eq:excess}. Controlling the expected absolute train--test gap would require an additional argument. For completeness, adding and subtracting the empirical optimum gives
\[
 \E[F(A)-F(w_F^\star)]
 =\E[F(A)-f(A)]+\E[f(A)-\inf f]+\E[\inf f-f(w_F^\star)] .
\]
The last term is nonpositive, which proves \eqref{eq:excess}. A stability estimate can be informative while the empirical optimization term remains large. The five comparisons below identify the term that the corresponding consistency argument actually controls.

%% file: appendix/a2-initialization.tex
\subsection{Relaxed initialization and the propagation of stored offsets}\label{app:init}
Write $h_i^t$ for the last stored endpoint before round $t$ and $d_t=m^{-1}\sum_i\norm{h_i^t-\w^t}^2$. The expected value of $d_t$ is $\Delta_t$ in Section~\ref{sec:init}. A stored endpoint need not come from round $t-1$. Under partial participation both its displacement from $\w^t$ and its age depend on the client's history.

\begin{lemma}[Historical and local-path contributions]\label{lem:init}
Let $U_t=(\w^t-\w^{t+1})/(\eta K)$ under \eqref{eq:ri}. Define $P_t=(mK)^{-1}\sum_{i,k}\E_t\norm{\w_{i,k}^t-\w^t}^2$. Then
\begin{align}
 b_t&=\frac{\beta}{m\eta K}\sum_i(h_i^t-\w^t)
      +\frac1{mK}\sum_{i,k}\E_t[\nabla f_i(\w_{i,k}^t)-\nabla f_i(\w^t)],\label{eq:explicit-init}\\
\norm{b_t}^2&\leq\frac{2\beta^2d_t}{\eta^2K^2}+2L^2P_t.\label{ap:init-bias}
\end{align}
\end{lemma}
\begin{proof}
Summing the $K$ local steps gives
$\w^t-\w_{i,K}^t=\beta(h_i^t-\w^t)+\eta\sum_{k<K}\g_{i,k}^t$.
Take the mean over a uniform cohort. Lemma~\ref{lem:cohort} and unbiased stochastic gradients along each potential trajectory give \eqref{eq:explicit-init} after subtracting $\nabla f(\w^t)$. Apply Jensen separately to each mean, local smoothness to the gradient differences, and $\norm{x+y}^2\le2\norm x^2+2\norm y^2$ to their sum.
\end{proof}
For a common initial model ($\beta=0$), Jensen on the single remaining term removes the factor two and recovers \eqref{eq:path}. With signed extrapolation the first term may cancel. A direct use of \eqref{ap:init-bias} without tracking this cancellation usually gives a loose rate.

\begin{lemma}[A local excursion estimate]\label{lem:excursion}
Let $D_k=m^{-1}\sum_i\E_t\norm{\w_{i,k}^t-\w^t}^2$ and $H_t=m^{-1}\sum_i\norm{\nabla f_i(\w^t)}^2$. Under conditional variance $\sigma_l^2$ and $4(\eta KL)^2\le1/2$,
\begin{equation}\label{ap:excursion}
 \max_{0\le k\le K} D_k
 \leq 4\beta^2 d_t+8\eta^2K^2 H_t+4\eta^2K^2\sigma_l^2 .
\end{equation}
\end{lemma}
\begin{proof}
Express the excursion after $k$ steps as its starting displacement plus $-\eta\sum_{j<k}g_{i,j}$. Jensen gives $\|\sum_{j<k}g_{i,j}\|^2\le k\sum_{j<k}\|g_{i,j}\|^2$. Unbiasedness and smoothness imply
$m^{-1}\sum_i\E_t\norm{g_{i,j}}^2\le2H_t+2L^2D_j+\sigma_l^2$.
Consequently $D_k\le2\beta^2d_t+2\eta^2K^2(2H_t+\sigma_l^2)+4\eta^2K^2L^2\max_{j\le K}D_j$. Take the maximum and absorb the last term. The bound also holds when $K=1$.
\end{proof}
This elementary bound records the origin of the three errors. The sharper noise dependence in Theorem~\ref{thm:init} uses conditional martingale variances of successive gradients before bounding the squared sums. The heterogeneity assumption gives $H_t\le G_h^2+B^2\norm{\nabla f(\w^t)}^2$.

\begin{lemma}[Cancellation for the cache used by the server]\label{lem:ri-cache}
For the actual last-seen endpoints define $\pi=s/m$, $\bar h=m^{-1}\sum_i h_i$, $h=\eta K$, and $\hat g_S=(sK)^{-1}\sum_{i\in S,k}g_{i,k}$. When $d_\pi=\pi-\beta(1-\pi)>0$, set $b_\pi=-\beta/d_\pi$, $\tilde h=h\pi/d_\pi$ and $y=w+b_\pi(\bar h-w)$. Then $y^+-y=-\tilde h\hat g_S$ for every drawn cohort and every set of local gradients. In particular
\begin{equation}\label{ap:init-cache}
 \norm{\E_t\hat g_S-\nabla f(y)}^2
 \le 2L^2\big(P_t+b_\pi^2d_t\big).
\end{equation}
\end{lemma}
\begin{proof}
Let $\bar h_S$ be the previous mean of the selected caches. The new selected caches are the endpoints whose average is $w^+$. Thus $\bar h^+-\bar h=\pi(w^+-\bar h_S)$, while \eqref{eq:ri} gives $w^+-w=-\beta(\bar h_S-w)-h\hat g_S$. Substituting both identities into $y^+-y$ gives a coefficient $-\beta[1-b_\pi(1-\pi)]-b_\pi\pi=0$ on $\bar h_S-w$. The gradient coefficient is $-h[1-b_\pi(1-\pi)]=-\tilde h$. For the bias, subtract the mean of $\nabla f_i(y)$ from the conditional expectation and apply Jensen, unbiasedness and smoothness. The squared distance from $w_{i,k}$ to $y$ is at most $2\norm{w_{i,k}-w}^2+2b_\pi^2\norm{\bar h-w}^2$. Another application of Jensen gives $\norm{\bar h-w}^2\le d_t$.
\end{proof}
This change of variables distinguishes a virtual all-client history centered at the latest average from a last-seen cache on an intermittently available client. Even if the initial cache mean equals the server model, refreshing only one cohort generally makes $\bar h^+\ne w^+$. The exact cancellation occurs at $y$, and the error in translating a gradient back to $w$ is at most $L|b_\pi|\sqrt{d_t}$.

\lead{A direct stationarity check for last-seen caches} Write $\langle V\rangle=T^{-1}\sum_{t<T}\E V_t$ for a time average in this paragraph. Let $A,B_s$ be the sampling constants obtained in the storage proof of Lemma~\ref{lem:smoo-track}. Its algebra applies to any potential new endpoints and reference equal to their active mean. Applying it with the old center $w$, then using Lemma~\ref{lem:excursion} for the new endpoints, gives
\[
 \E_t d_{t+1}\le(A+4 B_s\beta^2)d_t+B_s h^2(8H_t+4\sigma_l^2).
\]
For $\theta=A+4B_s\beta^2<1$, summation yields
$\langle d\rangle\le[d_0/T+B_s h^2(8\langle H\rangle+4\sigma_l^2)]/(1-\theta)$.
Let $V_t$ denote the conditional variance of $\hat g_S$. Jensen, unbiasedness and smoothness give $V_t\le2H_t+2L^2P_t+\sigma_l^2$. With $\tilde h\le1/L$, apply Lemma~\ref{lem:descent} at $y$. Use \eqref{ap:init-cache} to control the bias and $\norm{\nabla f(w)}^2\le2\norm{\nabla f(y)}^2+2L^2b_\pi^2d_t$ to translate back. With $D_y=f(y^0)-\inf f$,
\[
 \langle\norm{\nabla f(w)}^2\rangle
 \le \frac{4D_y}{\tilde h T}+4L^2\langle P\rangle
            +6L^2b_\pi^2\langle d\rangle+2L\tilde h\langle V\rangle .
\]
Lemma~\ref{lem:excursion} bounds $\langle P\rangle$ by $4\beta^2\langle d\rangle+h^2(8\langle H\rangle+4\sigma_l^2)$. Substitute these estimates and $\langle H\rangle\le G_h^2+B^2\langle\norm{\nabla f(w)}^2\rangle$. At fixed $\pi,\beta,L,B$ with $d_\pi>0$ and $\theta<1$, the coefficient multiplying the unknown stationarity average on the right is $O(h+h^2)$ as $h\downarrow0$. Taking $h$ small absorbs it. With seeded caches and $h$ of order $T^{-1/2}$ the resulting bound is $O(T^{-1/2})$. It uses conservative sampling and noise estimates. The sharper dependence on $s$ and $K$ in the published comparison below relies on its refined displacement estimate.

\lead{Proof of the source stationarity bound} We give the constants in the step that converts the source's excursion and sampling estimates into the rate. Write $G_t=\E\norm{\nabla f(\w^t)}^2$, $F_t=\E f(\w^t)$ and
\begin{align*}
 c_d&=\frac{10\beta^2L}{s(1-72\beta^2)},\quad
 k_1=17+\frac{1300\beta^2}{1-72\beta^2},\quad
 k_2=13+\frac{1020\beta^2}{1-72\beta^2},\\
 B_\eta&=\frac{\eta^2KL}{2s}\{k_2\sigma_l^2+3k_1K G_h^2\}.
\end{align*}
For the history in the published analysis, the refined one-round estimate in Appendix A.2 of \cite{sun2023understanding}, restated with the thesis notation, is
\begin{equation*}\label{ap:init-source}
 \frac{\lambda_0\eta K}{2}G_t
 \le F_t-F_{t+1}+c_d(\Delta_t-\Delta_{t+1})+B_\eta .
\end{equation*}
Its ingredients are smoothness at the broadcast, the local-gradient excursion in \eqref{eq:explicit-init}, and a recursion for the squared stored displacement on the active and inactive groups. The last recursion contributes the difference of successive $\Delta_t$. Its coefficient is finite for $72\beta^2<1$. The source takes a small enough step that the residual squared aggregate-gradient coefficient is nonpositive and chooses $\lambda_0>0$ below
\[
 1-\frac{39\eta KLB^2}{s}
   -\frac{3900\beta^2\eta KLB^2}{s(1-72\beta^2)}
   -12\eta^2K^2L^2B^2 .
\]
These conditions are the positive-descent qualification in Theorem~\ref{thm:init}. Summing \eqref{ap:init-source} eliminates both differences. The model-seeded caches give $\Delta_0=0$ and $F_T\ge\inf f$. Thus
\begin{equation*}\label{ap:init-finite}
 \frac1T\sum_{t<T}G_t\le
 \frac{2D_0}{\lambda_0\eta KT}
 +\frac{\eta L}{\lambda_0s}(k_2\sigma_l^2+3k_1K G_h^2).
\end{equation*}
This proves \eqref{eq:ri-bound} under the refined estimate. In the source calculation the mean historical offset makes no contribution to the expected descent inner product. For arbitrary last-seen caches that contribution is $-\beta\E\langle\nabla f(w),\bar h-w\rangle$ and needs control, as the direct argument above illustrates. With other initial caches the same calculation retains $2c_d\Delta_0/(\lambda_0\eta KT)$. 
For the Polyak--\L{}ojasiewicz (PL) comparison set $D_t=F_t-\inf f$, $r=1-\lambda_0\mu h$ with $0<r<1$. Substitution of $G_t\ge2\mu D_t$ in \eqref{ap:init-source} gives $D_{t+1}\le rD_t+c_d(\Delta_t-\Delta_{t+1})+B_\eta$. Iterating and summing the displacement differences by parts gives
\begin{align*}
 D_T&\le r^TD_0+B_\eta\frac{1-r^T}{1-r}
  +c_d\left[r^{T-1}\Delta_0-\Delta_T
      +(1-r)\sum_{j=1}^{T-1}r^{T-1-j}\Delta_j\right].\label{ap:ri-pl}
\end{align*}
The first term is at most $e^{-\lambda_0\mu hT}D_0$. The second is at most $\eta L(3k_1KG_h^2+k_2\sigma_l^2)/(2s\lambda_0\mu)$. The last sum is a geometrically weighted history. For common initialization without extrapolation, $c_d=0$. For $\beta>0$, the last-iterate rate reported in \cite{sun2023understanding} follows from this expression if the remaining weighted displacement is controlled at the claimed order. Seeded caches alone remove only $\Delta_0$. This is why the averaged stationarity estimate and a last-iterate PL guarantee have different requirements.

\lead{Stability of a relaxed start} Couple two executions on neighboring data. For the same sampled example the two parameter updates satisfy $\|u^+-v^+\|\le(1+\eta L)\|u-v\|$. At the changed example, add $\eta\|\nabla\ell(v;z)-\nabla\ell(v;z')\|$. Its conditional expected magnitude is at most $2\eta\upsilon$ whenever the paired gradient change is bounded by $2\upsilon$. For $G$-Lipschitz losses one may take $\upsilon=G$. In a mean over all potential clients, one replaced example contributes with weight $1/(mn)$ for equal client sizes $n$. Let $a_k^t$ denote the resulting mean expected distance. With a common step in the round, iterating gives
\begin{equation*}\label{ap:init-coupling}
 a_K^t\le\phi(t)a_0^t+\frac{2\upsilon}{mnL}(\phi(t)-1),
 \qquad \phi(t)=(1+\eta_tL)^K .
\end{equation*}
The local difference at entry is exactly $(1+\beta)(\w^t-\widetilde\w^t)-\beta(h_i^t-\widetilde h_i^t)$. This identity shows why the coupled cache must be tracked when clients return. It also explains why the sign of the extrapolation alone cannot prove an improvement in stability. Lemma~\ref{lem:replace} converts the resulting model comparison into a loss comparison.

The finite-horizon estimate in the later analysis \cite{shen2026rethinking} has the following form under its cached-state comparison, a bounded loss in $[0,U]$, loss Lipschitz constant $G$ and step $c/(t+1)$. For an admissible observation time $t_0$, put $a=cKL$:
\begin{equation*}\label{ap:init-cutoff}
 \varepsilon_G\le \frac{sUK}{mn}t_0+
 \frac{2\sigma_lG}{(1+2\beta)mnL}\left(\frac{T}{t_0}\right)^a .
\end{equation*}
The first term is a union bound for encountering the replaced record before $t_0$. The second propagates the later perturbation through the $K$ local steps. The balancing step in that proof assumes $1+2\beta\le\phi(t-1)/\phi(t)$ over the rounds to which it is applied. To see how this restriction enters, let $D_t'$ denote the coupled mean endpoint distance in that argument, $b=1+2\beta$ and $q_0=2\sigma_l/(mnL)$. Its cache comparison gives $D'_t\le b\phi(t)D'_{t-1}+q_0(\phi(t)-1)$. Put $v_t=(\phi(t)-1)/(b\phi(t)-1)$. Then $D'_t+q_0v_t\le b\phi(t)(D'_{t-1}+q_0v_t)$. When $b\phi(t)\le\phi(t-1)$, monotonicity gives $v_t\le v_{t-1}$ and the right side is at most $\phi(t-1)(D'_{t-1}+q_0v_{t-1})$. Multiplication from the no-hit cutoff, using $v_{t_0}\le1/b$, gives the late term from the product of the $\phi$'s. A harmonic step bounds that product by the corresponding power of the horizon-to-cutoff ratio, with endpoint shifts if the steps start at a different index. The early term covers the complementary trajectories. Balancing the two terms in \eqref{ap:init-cutoff} gives, up to the integer cutoff,
\[
 \varepsilon_G\le\frac{2}{mn}
 \left(\frac{2\sigma_lG}{(1+2\beta)L}\right)^{1/(1+a)}
 (sUKT)^{a/(1+a)} .
\]
For fixed $c,K,L$ the ratio $\phi(t-1)/\phi(t)$ tends to one. A fixed positive $\beta$ therefore needs an explicit finite-horizon check before using this comparison. The earlier analysis \cite{sun2023understanding} instead retains the trajectory divergence in its stability estimate. The empirical $\Delta_T$ in Table~\ref{tab:relaxation} is a single endpoint of that sequence.

%% file: appendix/a3-adaptivity.tex
\subsection{Adaptive directions, accumulated moments and a stationarity bound}\label{app:adaptive}
We distinguish the step length $h=K\eta_l\eta_g$ in the auxiliary sequence from the update length $\eta_l$ on a client. Write $G_g$ for the coordinate-wise bound on sampled gradients. If $S_t=\operatorname{diag}((\hat v^t)^{-1/2})$, every eigenvalue of the broadcast preconditioner is in $[s_-,s_+]$, with $s_-=1/\max\{G_g,\epsilon_v\}$ and $s_+=1/\epsilon_v$. Both endpoints matter. The upper bound controls fluctuations, and the lower bound measures descent along a nonzero global gradient.

\begin{lemma}[Historical cancellation and accumulation of the second moment]\label{lem:lada}
Suppose $h$ is constant over consecutive rounds and put $\bar a^t=(sK)^{-1}\sum_{i\in\cP_t,k}a_{i,k}^t$. For the auxiliary sequence in Theorem~\ref{thm:lada},
\begin{align}
 y^{t+1}-y^t&=-h\bar a^t,\label{ap:aux}\\
 \sum_{t<T}V_t&=\sum_j(\hat v_j^T-\hat v_j^0)\le dG_g^2,
 \quad V_t=\sum_j(\hat v_j^{t+1}-\hat v_j^t).\label{ap:accumulation}
\end{align}
\end{lemma}
\begin{proof}
Averaging \eqref{eq:lada} gives $\w^{t+1}-\w^t=-h\alpha\bar a^t+(1-\alpha)(\w^t-\w^{t-1})$. Subtract $(1-\alpha)(\w^t-\w^{t-1})$ and divide by $\alpha$. For the second identity, every active estimate starts at $\hat v^t$, then takes coordinate-wise maxima. Its terminal value stays between the starting value and $\max\{G_g^2,\epsilon_v^2\}$ because the uncorrected second moment is a convex combination of past squares. Averaging preserves the order. Sum the coordinate increments from the initial value $\epsilon_v^2$.
\end{proof}

\begin{lemma}[Freezing a denominator for one round]\label{lem:frozen}
Let $m_{i,k+1}$ denote the local first moment and $S_{i,k+1}$ the diagonal inverse-root second moment used in the same step. Suppress the round index on local variables. For the error
\[
 Q_t=\frac1{sK}\sum_{i\in\cP_t,k}(S_{i,k+1}-S_t)m_{i,k+1},
\]
one has $\|Q_t\|_1\le G_g V_t/(2\epsilon_v^3)$ and $\|Q_t\|^2\le G_g^2V_t/\epsilon_v^4$.
\end{lemma}
\begin{proof}
For $y\ge x\ge\epsilon_v^2$,
\[
 x^{-1/2}-y^{-1/2}=\frac{y-x}{\sqrt{xy}(\sqrt x+\sqrt y)}
 \le\frac{y-x}{2\epsilon_v^3},
 \qquad (x^{-1/2}-y^{-1/2})^2\le\frac{y-x}{\epsilon_v^4}.
\]
The second inequality uses $(\sqrt y-\sqrt x)^2\le y-x$. Multiply by $|m_{i,k+1,j}|\le G_g$ or its square. For the squared norm, apply Jensen to the client and step average first. Each coordinate increase is at most its active client's terminal increase. The mean of those terminal increases is exactly $\hat v^{t+1}-\hat v^t$. Summation over coordinates gives both assertions.
\end{proof}

\lead{Which noise is reduced by local work?} Starting the first moment at zero gives weights $\omega_k=(1-\beta_1^{K-k})/K$ and $c_K=\sum_{k=0}^{K-1}\omega_k$. Indeed $m_{i,j+1}=(1-\beta_1)\sum_{k\le j}\beta_1^{j-k}g_{i,k}$, and interchanging the order in $K^{-1}\sum_j m_{i,j+1}$ gives $\sum_k\omega_k g_{i,k}$. Thus $0<c_K\le1$, $\sum_k\omega_k^2\le1/K$. For $g=\nabla f(y^t)$, the average direction can be written exactly as
\begin{align*}
 \bar a^t &=c_K S_tg+N_t+P_t+Q_t,\label{ap:lada-decomp}\\
 N_t&=\frac{S_t}{s}\sum_{i\in\cP_t,k}\omega_k[g_{i,k}-\nabla f_i(w_{i,k})]
     +c_KS_t\left(\frac1s\sum_{i\in\cP_t}\nabla f_i(y^t)-g\right),\nonumber\\
 P_t&=\frac{S_t}{s}\sum_{i\in\cP_t,k}\omega_k[\nabla f_i(w_{i,k})-\nabla f_i(y^t)].\nonumber
\end{align*}
The term $P_t$ uses local gradients along the amended path. The term $Q_t$ accounts for the change in the denominator caused by the same samples as the numerator. Freezing the broadcast denominator before conditioning prevents an invalid independence assumption.

Assume independent client random streams, unbiased local gradient noise of conditional variance at most $\sigma_l^2$, and population gradient dispersion bounded by $\sigma_g^2$. The local noises are martingale differences and the second term in $N_t$ is the sample mean of a fixed population at $y^t$. Conditional on the cohort the local-noise sum has mean zero. Applying Lemma~\ref{lem:cohort} gives
\begin{equation*}\label{ap:adaptive-noise}
 \E_t N_t=0,\qquad
 \E_t\norm{N_t}^2\le V_N
 :=s_+^2\left(\frac{\sigma_l^2}{sK}+c_K^2\chi_s\sigma_g^2\right).
\end{equation*}
This elementary estimate displays the distinct $K$-dependence of within-client and active-cohort noise. Put $A=\sqrt d\,G_g/\epsilon_v$. Both a local adaptive direction and the historical direction have norm at most $A$, the latter by induction in \eqref{eq:lada} from $q^0=0$. Therefore
\begin{equation*}\label{ap:adaptive-path}
 \norm{w_{i,k}-y^t}\le D=A\!\left(K\eta_l+\frac{1-\alpha}{\alpha}h\right),
 \qquad \norm{P_t}\le D_P=c_Ks_+LD .
\end{equation*}
The first summand bounds travel from the broadcast. The second bounds the displacement from the broadcast to the auxiliary iterate.

\lead{A direct finite-horizon calculation} By \eqref{ap:aux} and smoothness, the one-round cost is at most $-h\inner{g}{\E_t\bar a^t}+Lh^2\E_t\norm{\bar a^t}^2/2$. In its linear term gives descent of at least $hc_Ks_-\norm g^2$. Bound $|\inner g{P_t}|$ by $c_Ks_-\norm g^2/4+D_P^2/(c_Ks_-)$, and $|\inner g{Q_t}|$ by $G_g\|Q_t\|_1$. In its quadratic term, the square of a sum of four vectors is at most four times the sum of their squares. If $h\le s_-/(8Lc_Ks_+^2)$, the contribution $2Lh^2c_K^2s_+^2\norm g^2$ consumes at most one quarter of the descent. Summing the result and using Lemma~\ref{lem:frozen} gives the explicit estimate
\begin{align*}
 \frac{c_Ks_-}{2T}\sum_{t<T}\E\norm{\nabla f(y^t)}^2\le \frac{f(y^0)-\inf f}{hT}+\left(\frac1{c_Ks_-}+2Lh\right)D_P^2+2LhV_N + \left(\frac{G_g^2}{2\epsilon_v^3}+\frac{2LhG_g^2}{\epsilon_v^4}\right)\frac{d G_g^2}{T}.
\end{align*}
This bound proves descent for the same auxiliary iterate using only the displayed assumptions. It deliberately retains the sampling variance and a worst-case local path. Its dependence on $K$ should be read from \eqref{ap:adaptive-noise} and \eqref{ap:adaptive-path}.

\lead{Reduction to the published rate} The tighter path and squared-update estimates in the FedLADA analysis \cite{sun2023efficient,sunthesis} give the following grouped form before choosing the steps. Set $G_\vartheta=\max_{i,k,t}\|S_{i,k}^t\|_2$ and $C=(1-\beta_1/((1-\beta_1)K))/2>0$. With the source's local-path coefficients $C_2,C_3$ (both proportional to $K$ for fixed hyperparameters), an upper bound $G_c$ on $\E\|q^t\|^2$, positive tuning constant $v$, and $\sigma^2=\sigma_l^2+\sigma_g^2+\sup_w\|\nabla f(w)\|^2$, its estimate reads
\begin{align*}
 \frac1T\sum_{t<T}\E\norm{\nabla f(y^t)}^2
 &\le \frac{f(y^0)-\inf f}{h C G_\vartheta T}
 +\frac{2h\beta_1^2G_\vartheta L\sigma_l^2}{CsK}
 +\frac{dG_g^4}{CG_\vartheta T}\left(\frac{hL}{\epsilon_v^4}
                                     +\frac1{2\epsilon_v^3}\right) \nonumber\\
 &\quad+\frac{(1+v)L^2C_3\eta_l^2\sigma^2}{2C}
 +\frac{2L^2G_c}{C}\left((1+v) C_2\eta_l^2+
 (1+1/v)\left(\frac{1-\alpha}{\alpha}\right)^2 h^2\right).
 \label{ap:lada-source}
\end{align*}
For completeness, one choice of the local-path constants in the thesis is
\[
 C_1=\frac{\alpha(1+b)(1+1/b)}{1+b-\beta_1},\quad
 C_2=6(1-\alpha)(1+1/b)KG_\vartheta^2,\quad
 C_3=24(1-\beta_1)KG_\vartheta^2 C_1 ,
\]
where $b>0$ is the inner Young-inequality parameter. A bound used for the historical direction is $G_c=dG_g^4/\epsilon_v^4+\beta_1^2G_\vartheta^2\sigma_l^2/(sK)+2G_\vartheta^2(\sigma_g^2+\sup_w\norm{\nabla f(w)}^2)$. The first line comes from telescoping $f(y^t)$, the local martingale term and the denominator increments. The second line retains local drift and the historical shift in $y^t$. In the source proof nonpositive squared-direction terms are discarded under
\[
 h\le\min\left\{\frac1{4L},\frac{\beta_1s(m-1)(1+v)}{32(m-s)LG_\vartheta^3}\right\},
 \qquad
 \eta_l\le\frac1{2\sqrt{2\alpha(1-\beta_1)K}\,G_\vartheta L}
\]
for $1<s<m$. Choosing $h\asymp\sqrt{sK/T}$ and $\eta_l\asymp (KT)^{-1/2}$ balances $1/(hT)$ with $h/(sK)$. The other terms are proportional to $h/T$, $1/T$, $C_3\eta_l^2$, $C_2\eta_l^2$ and $h^2$. Substitution gives \eqref{eq:lada-rate} with the source's parameter-dependent constants. For example, $h^2\asymp sK/T$ and $C_3\eta_l^2=O(1/T)$ for fixed remaining hyperparameters. A speedup statement that varies $s$ or $K$ with $T$ must retain these terms and recheck both step constraints. Neither calculation supplies a bound on wall-clock time.

\begin{lemma}[Fixed point in the controlled scaling experiment]\label{lem:synthetic}
In \eqref{eq:control-dir} let $\bar H=m^{-1}\sum_iH_i$, $\bar a=m^{-1}\sum_i a_i$ and $\overline{Ha}=m^{-1}\sum_i H_i a_i$. With positive diagonal curvatures, define
\[
 B_\gamma=(1-\gamma)I+\gamma\bar P\bar H,\qquad
 b_\gamma=(1-\gamma)\bar a+\gamma\bar P\,\overline{Ha}.
\]
Then the aggregate update has unique fixed point $w_\gamma=B_\gamma^{-1}b_\gamma$. After $t$ constant steps,
\begin{equation}\label{ap:quadratic-trajectory}
 w^t-w_\gamma=(I-\eta B_\gamma)^t(w^0-w_\gamma),\qquad
 u_\gamma(w^\star)=(1-\gamma)(w^\star-\bar a).
\end{equation}
\end{lemma}
\begin{proof}
The exact local gradient is $H_i(w-a_i)$ and $P_iH_i=I$. Expanding the weighted direction gives $u_\gamma(w)=B_\gamma w-b_\gamma$. Every diagonal entry of $B_\gamma$ is positive. Subtract its zero from both sides of a gradient step to get the recurrence and iterate. Evaluating at the global optimum, where $\bar H w^\star=\overline{Ha}$, gives the second identity.
\end{proof}
The factor $1-\gamma$ predicts the residual at the optimum for every individual problem instance. The remaining objective gap is a different measurement, $f(w^t)-f(w^\star)=\tfrac12(w^t-w^\star)^\top\bar H(w^t-w^\star)$. It need not vary linearly with $\gamma$. If $0<\eta<2/\lambda_{\max}(B_\gamma)$, the transient in \eqref{ap:quadratic-trajectory} decays on every coordinate but the distance from $w_\gamma$ to $w^\star$ can persist. For common curvatures, $w^\star=\bar a$ and all five scalings have the same fixed point. These identities give an independent check on the trajectories in Figure~\ref{fig:scaling-paths}. The experimental $\gamma$ mixes prescribed matrices, while the learned FedLADA amendment $\alpha$ mixes update directions.

%% file: appendix/a4-sharpness.tex
\subsection{Shared perturbations, model tracking and margins}\label{app:sharp}
The perturbation radius is common to both objectives in \eqref{eq:sharp-obj}. Write $g_i=\nabla f_i(w)$, $g=\sum_i p_i g_i$, and take a closed Euclidean ball. Its compactness and continuity of the losses ensure the maxima exist.

\begin{lemma}[Uniform first-order expansion on a ball]\label{lem:sharp-linear}
For any differentiable $L$-smooth $h$ on $w+B_\rho$,
\[
 \left|\max_{\|u\|\le\rho}h(w+u)-h(w)-\rho\norm{\nabla h(w)}\right|
 \le L\rho^2/2.
\]
\end{lemma}
\begin{proof}
Integrating $\nabla h(w+tu)-\nabla h(w)$ along the segment bounds the absolute Taylor remainder by $L\norm u^2/2$. The maximum of the linear term is $\rho\norm{\nabla h(w)}$ by Cauchy--Schwarz. Evaluating the lower Taylor bound at a maximizer of this linear term and maximizing the upper Taylor bound prove the two inequalities, including when the gradient is zero.
\end{proof}
\lead{Proof of Proposition~\ref{prop:sharp}} Apply the lemma to $f$ and to each $f_i$, average the latter bounds and subtract. The two remainders total at most $L\rho^2$, giving \eqref{eq:sharp-gap}. For a maximizing shared perturbation $u^\star$, the exact gap equals $\sum_i p_i(\max_{u_i}f_i(w+u_i)-f_i(w+u^\star))$. Every summand is nonnegative, so the gap vanishes exactly when $u^\star$ maximizes every positive-weight term. To first order the gap vanishes whenever the nonzero $g_i$ point along a common ray, which is exactly the equality condition for the triangle inequality. For the two equal-weight losses $(w\pm1)^2/2$ at $w=0$, the first-order gap is $\rho$ and the second-order terms cancel exactly. A discrepancy can therefore be present even when the client models coincide.

\lead{Exact-solve tracking} We now follow the virtual-sequence construction in the FedSMOO analysis \cite{sun2023dynamic,sunthesis}. At the start of a round let $x_i$ be the latest stored model, $\ell_i$ the model multiplier with the gradient-tracking sign, $z=w+b\bar\ell$ the virtual mean, and $b=\beta_s$. For each potential endpoint $v_i$, let $s_i$ be its perturbation, with $\norm{s_i}\le\rho$. The optimality and multiplier updates obey
\begin{equation*}\label{ap:smoo-implicit}
 v_i=w+b\{\ell_i-\nabla f_i(v_i+s_i)\},\qquad
 \ell_i^{+}=\ell_i-(v_i-w)/b=\nabla f_i(v_i+s_i)
 \quad(i\in S).
\end{equation*}
The stored pair of an inactive client is carried forward. A warm history means $\ell_i=\nabla f_i(x_i+s_i^{\rm old})$ with $\norm{s_i^{\rm old}}\le\rho$, a relation subsequently preserved by \eqref{ap:smoo-implicit}. An arbitrary initial pair would require an extra tracking residual in the first round. The next virtual mean is $z^+=s^{-1}\sum_{i\in S}v_i$. Uniform sampling yields $\E_t(z^+-z)=-b\,m^{-1}\sum_i\E_t\nabla f_i(v_i+s_i)$. No missing client has to compute a virtual endpoint in the implementation.

\begin{lemma}[One-round perturbation and storage bounds]\label{lem:smoo-track}
Let $G_t=\E\norm{\nabla f(z)}^2$, $\nu_t=m^{-1}\sum_i\E\norm{v_i-z}^2$ and $\delta_{t-1}=m^{-1}\sum_i\E\norm{x_i-z}^2$. With $F_t=\E f(z)$,
\begin{align}
 F_{t+1}-F_t &\le -bG_t/2+(L/2+bL^2)\nu_t+bL^2\rho^2, \label{ap:smoo-desc}\\
 (1-12b^2L^2)\nu_t&\le 36b^2L^2\delta_{t-1}+12b^2G_t+30b^2L^2\rho^2.
 \label{ap:smoo-track}
\end{align}
There exist $A<1$ and $B<\infty$, depending on the sampling fraction, such that $\delta_t\le A\delta_{t-1}+B\nu_t$.
\end{lemma}
\begin{proof}
Jensen's inequality gives $\E\norm{z^+-z}^2\le\nu_t$. The mean perturbed gradient differs from $\nabla f(z)$ by a vector whose expected squared norm is at most $2L^2(\nu_t+\rho^2)$. Apply smooth descent at $z$ to $z^+-z$ and use $-2\inner uv\le-\norm u^2+\norm{u-v}^2$ in the linear term. This proves \eqref{ap:smoo-desc}. More specifically, replacing a designated shared perturbation $s_0$ by $s_i$ in a local gradient costs at most $L\norm{s_i-s_0}$ at the same model. The radius-only estimate above permits all directions in the ball.

For the second inequality set $g_i^0=\nabla f_i(z)$, $r_i=\ell_i-g_i^0$ and $h_i=\nabla f_i(v_i+s_i)-g_i^0$. Equation~\eqref{ap:smoo-implicit} becomes $v_i-z=b[(r_i-\bar r)-\nabla f(z)-h_i]$. Warm tracking and smoothness give
\[
 m^{-1}\sum_i\E\norm{r_i}^2\le 2L^2(\delta_{t-1}+\rho^2),\qquad
 m^{-1}\sum_i\E\norm{h_i}^2\le 2L^2(\nu_t+\rho^2).
\]
Centering decreases the first squared norm. Bounding the squared norm of three vectors by three times the sum of their squares gives
$\nu_t\le 6b^2L^2(\delta_{t-1}+\nu_t)+3b^2G_t+12b^2L^2\rho^2$.
Enlarging these coefficients gives the form \eqref{ap:smoo-track} used in the thesis.

For an explicit storage estimate let $\pi=s/m$, $c=\pi/(2-\pi)$ and $A=(1-\pi)(1+c)<1$. Conditional on all potential endpoints, the active part of the new storage error is
$m^{-1}\sum_{i\in S}\norm{v_i-z}^2-\pi\norm{z^+-z}^2$.
For each inactive client use $\norm{x_i-z^+}^2\le(1+c)\norm{x_i-z}^2+(1+1/c)\norm{z^+-z}^2$. Take expectations and use the bound on the mean displacement. The claimed estimate holds with
\[
 B=\pi+\max\{0,(1-\pi)(1+1/c)-\pi\}.
\]
The source analysis gives a sharper storage coefficient for its auxiliary sequences. This explicit choice suffices for the order in Theorem~\ref{thm:smoo} and retains the dependence on participation.
\end{proof}

\lead{Proof of Theorem~\ref{thm:smoo}} Set $C_f=L/2+bL^2$. Choose $b>0$ sufficiently small and define
\begin{equation*}\label{ap:smoo-potential}
 p=\frac{C_f}{1-12b^2L^2-36b^2L^2B/(1-A)},\qquad
 q=\frac{36pb^2L^2}{1-A},\qquad \zeta=\tfrac12-12pb .
\end{equation*}
The denominator tends to one and $\zeta$ to $1/2$ as $b$ decreases for fixed participation. Add $q(\delta_t-\delta_{t-1})$ to \eqref{ap:smoo-desc}. Since $C_f+qB=p(1-12b^2L^2)$, substitute \eqref{ap:smoo-track}. The coefficient of $\delta_{t-1}$ becomes $36pb^2L^2-q(1-A)=0$. Thus
\[
 b\zeta G_t\le (F_t+q\delta_{t-1})-(F_{t+1}+q\delta_t)
                   +b(1+30pb)L^2\rho^2 .
\]
Summing and using $F_{T+1}\ge\inf f$ and $\delta_T\ge0$ proves \eqref{eq:smoo-rate}. If $\rho^2=O(1/(sT))$ this residual also decays with $T$. At fixed $\rho$ the displayed estimate retains the perturbation term. Replacing an exact solve by a finite sequence of minibatch updates would add its first-order residual to \eqref{ap:smoo-implicit}.

\lead{From a single predictor to a margin comparison} Consider the deployed, globally perturbed network $h_W$, whose perturbation is already included in $W$. The random $U_j$ in this argument are separate analytical perturbations of each layer. Suppose the bias-free network has $D$ layers, input norm at most $R$, and 1-Lipschitz activations vanishing at zero. Write $P=\prod_j\norm{W_j}_2$ and $r_j=\norm{U_j}_2/\norm{W_j}_2$ for nonzero layers. Replacing layers one at a time and applying the Lipschitz bound to each remaining composition gives
\begin{equation*}\label{ap:layer-perturb}
 \sup_{\norm x\le R}\norm{h_{W+U}(x)-h_W(x)}
 \le RP\sum_{j=1}^D r_j\prod_{k<j}(1+r_k)
 \le RP\,e^{\sum_j r_j}\sum_j r_j .
\end{equation*}
In the telescoping product the first $j-1$ layers may already be perturbed and the later layers retain their original norms. This explains why all layer norms enter the result.

Use the spectral PAC--Bayes margin lemma with its sampling and fixed-prior conditions \cite{neyshabur2018a,farnia2018generalizable,sun2023dynamic}. Its input is a distribution on $U$ giving probability at least $1/2$ to a uniform logit change at most $\gamma/4$. On that event a true-class margin changes by at most $\gamma/2$, since it is a difference of two logits. For independent isotropic Gaussian matrices with entry scales $\sigma_j$, a Gaussian spectral-norm tail and a union bound yield $\norm{U_j}_2\le C\sigma_j\sqrt{d\log(4dD)}$ simultaneously with the required probability. In the nontrivial margin regime one may take scales of order
\[
 \sigma_j=\frac{c_0\gamma\norm{W_j}_2}{D RP\sqrt{d\log(4dD)}}
\]
for a small numerical $c_0$. Equation~\eqref{ap:layer-perturb} then gives the desired logit change. The prior scales must be chosen independently of the training sample. For the usual positively homogeneous normalization one can equalize the layer norms without changing the function, discretize the common scale in advance and charge its logarithmic selection cost as in the cited lemma. Arbitrary data-dependent Gaussian priors do not justify this step.

The Gaussian shift cost relative to a centered prior of the same scale is $\tfrac12\sum_j\norm{W_j}_F^2/\sigma_j^2$. Substituting the scales gives order $D^2R^2d\log(4dD)V(W)/\gamma^2$. Applying the cited margin lemma to the $mn$ independent training examples under its population-mixture sampling conditions adds the logarithmic confidence term. For $0<\gamma\le1$ this gives \eqref{eq:margin}, with numerical constants absorbed in the order notation. It concerns a single set of logits on every example. Independently perturbed client classifiers need not share either the margins or the layer norms used in this argument.

\lead{Why a stratified sample can be used in this step} Let $N=mn$ and fix any perturbed predictor $h$. For a margin threshold $\gamma/2$, $R(h)$ is its mean error over the uniform population mixture and $\widehat R(h)$ is the mean of the $N$ independent bounded indicators, $n$ from each client. Their expectations average to $R(h)$. Hoeffding's inequality gives $\Pr(N(\widehat R(h)-R(h))^2>x)\le 2e^{-2x}$, even if the client distributions differ. Integration yields $\E_{\rm sample}\exp[N(\widehat R(h)-R(h))^2]\le3$. For a data-independent prior $P_0$, average over $h\sim P_0$ and apply Markov's inequality. On one event of probability at least $1-\delta$, a change of measure and Jensen give, simultaneously for all posteriors $Q$,
\[
 R(Q)-\widehat R(Q)
 \le \sqrt{\{\mathrm{KL}(Q\Vert P_0)+\log(3/\delta)\}/N}.
\]
Here the change of measure uses $\E_Q X\le\mathrm{KL}(Q\Vert P_0)+\log\E_{P_0}e^X$, which follows by writing the integral under $Q$ and applying Jensen to the logarithm. Independence of training examples and agreement between sampling weights and the target mixture are the sampling requirements used in this step.

Let $E$ denote the uniform logit-change event above and apply the displayed inequality to the Gaussian posterior conditioned on $E$. With $Q(E)\ge1/2$, the entropy identity for a distribution split on an event gives $\mathrm{KL}(Q(\cdot\mid E)\Vert P_0)\le2[\mathrm{KL}(Q\Vert P_0)+\log 2]$. For every draw in $E$, a zero-margin error of $W$ implies a $\gamma/2$-margin error of the perturbed network. Any such empirical error implies a $\gamma$-margin error of $W$. Substitution gives the deterministic predictor comparison used in \eqref{eq:margin}. A predeclared scale grid contributes its logarithmic cardinality through a union bound or a prior mixture. This argument specifies the predictor, the client weights and the cost of selecting its norm scale.

%% file: appendix/a5-duality.tex
\subsection{Primal--dual tracking under partial participation}\label{app:dual}
Write $z=\bar w^t$, $w=z+\bar\lambda/\tau$, and $z^+=s^{-1}\sum_{i\in S}x_i$ for the next virtual mean. The server broadcast $w$ and the mean $z$ are distinct. Fix the current state and draw all potential local solves independently of selection. For a potential endpoint $x_i$, put $e_i=\nabla f_i(x_i)+\lambda_i+\tau(x_i-w)$. We assume $\E_t e_i=0$, $\E_t\norm{e_i}^2\le\epsilon$ with $\epsilon=\varepsilon_{\rm loc}$, and $L$-smooth $f_i$. If $z_i$ is the endpoint used in the multiplier update \eqref{eq:dual}, direct cancellation gives
\begin{equation*}\label{ap:dual-res}
 \nabla f_i(x_i)+\lambda_i^+=e_i+\tau(z_i-x_i).
\end{equation*}
For an active client only the local-solver residual remains. For an absent client the second term is the error made by imputing $z^+$ for its potential endpoint. This identity explains why a common server increment restores only part of the stationary relation.

\begin{lemma}[Displacement and descent at the virtual mean]\label{lem:dual-desc}
Let $d_i=x_i-z$, $l_i=\lambda_i-\bar\lambda$, $g=\nabla f(z)$, $r_i=l_i+\nabla f_i(z)-g$, and $M=m^{-1}\sum_i\norm{r_i}^2$. Thus $\bar r=0$. With $R=m^{-1}\sum_i\E_t\norm{d_i}^2$, $D_\tau=(\tau-L)^2$ and $c=L(1+L/\tau)/2$, for $\tau>L$,
\begin{equation}\label{ap:dual-desc}
 R\le\frac{\norm g^2+M+\epsilon}{D_\tau},\qquad
 \E_t f(z^+)-f(z)\le-\frac{\norm g^2}{2\tau}+cR .
\end{equation}
\end{lemma}
\begin{proof}
Let $h_i=\nabla f_i(x_i)-\nabla f_i(z)$ and $\xi_i=e_i-h_i$. Local optimality with residual gives $\tau d_i=-g-r_i+\xi_i$. Because $\norm{h_i}\le L\norm{d_i}$, $(\tau-L)\norm{d_i}\le\norm{e_i-g-r_i}$. Square before averaging. The cross terms with $e_i$ disappear conditionally, and $m^{-1}\sum_i\norm{g+r_i}^2=\norm g^2+M$. This proves the first claim. Uniform sampling gives $\tau\E_t(z^+-z)=-g-m^{-1}\sum_i\E_th_i$, and Jensen gives $\E_t\norm{z^+-z}^2\le R$. In the smooth upper quadratic for $f$, bound the cross term by $L\norm g\sqrt R/\tau\le(\norm g^2+L^2R)/(2\tau)$. The second claim follows.
\end{proof}

\lead{Reduction using the displacement estimates in the thesis} With total expectations let $R_t=\E R$, $F_t=\E f(z)$ and $G_t=\E\norm g^2$. The virtual-history comparison in the A-FedPD proof \cite{sun2024fedpd,sunthesis} supplies the additional input
\begin{equation*}\label{ap:dual-source}
 (1-4L^2/\tau^2)R_t
 \le (32L^2/\tau^2)R_{t-1}+(4L^2/\tau^2)G_t+4\epsilon/\tau^2 .
\end{equation*}
In this comparison the disagreement of a virtual endpoint from the new virtual mean is bounded by $4R_t$, by inserting $z$ and applying Jensen to the active mean. Controlling the previous virtual endpoint used in the gradient history is an additional displacement estimate. A frozen or imputed multiplier in \eqref{ap:dual-res} need not equal the negative gradient at that endpoint.

\lead{Proof of Theorem~\ref{thm:dual} under \eqref{ap:dual-source}} Put $a_0=1-4L^2/\tau^2$, $b_0=32L^2/\tau^2$ and $v=c/(a_0-b_0)$. For $\tau>6L$, $a_0>b_0$. Then take $\tau$ large enough that $a_\tau=1/(2\tau)-4vL^2/\tau^2>0$. Add $vb_0(R_t-R_{t-1})$ to the objective descent in Lemma~\ref{lem:dual-desc}. The coefficient of $R_t$ on the right becomes $c+vb_0=va_0$, so \eqref{ap:dual-source} yields
\[
 a_\tau G_t\le (F_t+vb_0R_{t-1})-(F_{t+1}+vb_0R_t)+4v\epsilon/\tau^2 .
\]
Telescoping proves \eqref{eq:dual-rate} with $c_\tau=vb_0$ and residual $4v\epsilon/(a_\tau\tau^2)$. The following direct argument isolates the participation condition for the actual centered multipliers. It also identifies a case in which the displacement input cannot be inferred from smoothness and exact local solves alone.

\begin{lemma}[Contraction of the centered multiplier error]\label{lem:dual-center}
For a uniform cohort with $2\le s\le m$, set $\kappa=(s-1)/(m-1)$ and $C_\kappa=4+2/\kappa$. Let $M^+$ evaluate $M$ at the updated multipliers and $z^+$. Then
\begin{equation}\label{ap:dual-center}
 \E_t M^+\le(1-\kappa/2)M+3\epsilon+C_\kappa L^2R .
\end{equation}
\end{lemma}
\begin{proof}
Use the client-space inner product $\inner uv_{\rm av}=m^{-1}\sum_i\inner{u_i}{v_i}$. Let $(P_Sd)_i=d_i-\bar d_S$ on $S$, zero otherwise. This is the orthogonal projection onto vectors supported on $S$ with sum zero. Since $\lambda_i^+-\bar\lambda^+=l_i+\tau(P_Sd)_i$, the centered residual after the multiplier update, still evaluated at the old $z$, is $(I-P_S)r+P_S\xi$. The two terms are orthogonal for each cohort and realization. By Lemma~\ref{lem:cohort},
\[
 \E_t\norm{P_Sr}_{\rm av}^2
 =\frac{s}{m}\left(M-\E_t\norm{\bar r_S}^2\right)
 =\frac{s-1}{m-1}M .
\]
Furthermore $\E_t\norm{P_S\xi}_{\rm av}^2\le2\epsilon+2L^2R$. Moving the center to $z^+$ adds a centered change of local gradients of squared client norm at most $L^2\norm{z^+-z}^2$. Apply $\norm{u+v}^2\le(1+\kappa/2)\norm u^2+(1+2/\kappa)\norm v^2$. Since $(1+\kappa/2)(1-\kappa)\le1-\kappa/2$, the stated coefficients follow.
\end{proof}

\lead{A bound directly for the server update} Combine Lemmas~\ref{lem:dual-desc} and \ref{lem:dual-center} with $a=4c/(\kappa D_\tau)$ and $u=(c+aC_\kappa L^2)/D_\tau$. A sufficient penalty choice is $D_\tau\ge\max\{4C_\kappa L^2/\kappa,8\tau c\}$. It ensures $u\le 2c/D_\tau\le\min\{a\kappa/2,1/(4\tau)\}$. For $V=f(z)+aM$ the coefficient of $M$ in $\E_t V^+-V$ is then nonpositive, and
\[
 \E_t V^+-V\le-\norm g^2/(4\tau)+(u+3a)\epsilon .
\]
For an initial objective gap $D_f$ and centered error $M_1$ this yields
\begin{equation*}\label{ap:dual-direct}
 \frac1T\sum_{t=1}^T\E\norm{\nabla f(z^t)}^2
 \le\frac{4\tau(D_f+a\E M_1)}{T}+4\tau(u+3a)\epsilon .
\end{equation*}
This derivation obtains the same stationarity order without identifying the imputed multiplier with a previous local gradient. The sufficient penalty deteriorates for small cohorts. Even with perfect solves, heterogeneity can make $\bar x_S$ an imperfect surrogate for every individual absent client.\\

\lead{Boundary at one active client} If $m=2$, $s=1$ and both initial multipliers are equal, $P_S=0$ and they remain equal. Take $f_i(x)=(x-a_i)^2/2$, $a_1=-1$, $a_2=1$, independent uniform selection in each round, and exact solves. Equation~\eqref{eq:dual} gives $z^+=(\tau z+a_{I_t})/(1+\tau)$. The cross term in its squared expectation is zero. Hence $\E(z^+)^2=[\tau^2\E z^2+1]/(1+\tau)^2$ and $\E|\nabla f(z^t)|^2\to1/(1+2\tau)$. At fixed $\tau$ uniform selection alone cannot imply a vanishing stationarity average for this one-client regime.

\lead{Local stability and the exposure to a replaced example} Couple the local steps with identical random draws on neighboring data sets. With $v=u-w$ the update is $v^+=(1-\eta\tau)v-\eta(\nabla\ell(w+v)+\lambda_i)$. At a shared sample, for $0\le\eta\tau\le1$ and $\tau>L$, the distance in $v$ is multiplied by at most $1-\eta(\tau-L)$, plus $\eta(L\norm{\Delta w}+\norm{\Delta\lambda_i})$ due to the other states. If the draw is the replaced example, its gradient change at the same point adds at most $2\eta G$. Geometric summation over $K$ steps therefore controls inherited disagreement even at a long local interval. The two server-side states still require a coupled estimate.

The source uses a nonnegative potential $H_t$ that dominates the expected model distance and also includes the multiplier distance \cite{sun2024fedpd,sunthesis}. Its round estimate has the form $H_{t+1}\le(1+c_0\eta_t KL)H_t+2c_0\eta_t KG/(mn)$. For $\eta_t=\eta_0/(t+1)$ and $c_0\eta_0K\le\mu$, add $2G/(mnL)$ to both sides. Starting from identical states at round $r\ge1$, multiplication gives
\[
 H_T\le\frac{2G}{mnL}\big[(T/r)^{\mu L}-1\big].
\]
Indeed, $\log(1+x)\le x$ and $\sum_{t=r}^{T-1}(t+1)^{-1}\le\log(T/r)$. The early-exposure term in Lemma~\ref{lem:replace} depends on how many draws the cutoff counts. Under the source's exposure estimate $sr/(mn)$, balancing $Hsr/(mn)$ (where $H$ bounds the loss) against $2G^2(T/r)^{\mu L}/(mnL)$ yields the expression in Section~\ref{sec:dual}, up to rounding. If instead all $K$ independent within-client draws in a round may encounter the replacement, the round exposure is at most $(s/m)[1-(1-1/n)^K]\le sK/(mn)$. With that accounting the first term has an additional factor up to $K$. A schedule with $\eta_0=O(1/K)$ controls amplification after a hit. The sampling rule determines the probability of the first hit.

%% file: appendix/a6-topology.tex
\subsection{Network mixing, sample exposure and the stability comparison}\label{app:network}
Two mechanisms appear when the coordinator is replaced by a graph. Mixing preserves the average over clients but redistributes their deviations. In a neighboring-data argument, changing the participant set also changes the probability that an altered example is seen. We keep the two calculations separate. The sample size is $n$ per client, $N=TK$ is the number of local step indices, and $s=m_p$ is the centralized cohort size.

\begin{lemma}[Mixing with fresh local increments]\label{lem:mix}
Let $A$ be symmetric, nonnegative and doubly stochastic, $J=\mathbf1\mathbf1^\top/m$ and $\lambda=\norm{A-J}_2<1$. For the iteration $X^{t+1}=A^qX^t+D^t$, write $u_t=\norm{(I-J)D^t}_F$. Then
\begin{equation}\label{ap:forced}
 \norm{(I-J)X^t}_F
 \le\lambda^{qt}\norm{(I-J)X^0}_F+
       \sum_{j=0}^{t-1}\lambda^{q(t-1-j)}u_j .
\end{equation}
\end{lemma}
\begin{proof}
Since $J^2=J$ and $AJ=JA=J$, we have $A^q=J+(A-J)^q$ for $q\ge1$. The disagreement after a mixing step is $(A-J)^q(I-J)X^t$ and its Frobenius norm is at most $\lambda^q\norm{(I-J)X^t}_F$. Adding the new increment and iterating proves the claim. If increments precede mixing, the new input is itself multiplied by $A^q$ and its disagreement gains an additional factor $\lambda^q$.
\end{proof}
Setting all $D^t$ to zero gives \eqref{eq:gossip} and the exchange budget in Section~\ref{sec:topology}. For a uniform envelope $u_t\le u$, the forced part is at most $u/(1-\lambda^q)$. Even from identical initial models a new local step can create disagreement. Moreover, for local $L$-smooth losses at the rows of $X$,
\[
 \left\|\frac1m\sum_i\nabla f_i(X_i)-\nabla f(\bar X)\right\|
 \le\frac{L}{m}\sum_i\norm{X_i-\bar X}
 \le\frac{L}{\sqrt m}\norm{(I-J)X}_F .
\]
The separate terms distinguish conserving the model mean from evaluating the gradient at that mean.

\begin{lemma}[A uniform bound on the mixing sum]\label{lem:kernel}
Let $0<\lambda<1$, $0\le\nu<1$. For $t\ge1$ define $K_t=t^\nu\sum_{j=0}^{t-1}\lambda^{t-1-j}(j+1)^{-\nu}$. At $\nu=0$, $K_t\le1/(1-\lambda)$. For $0<\nu<1$,
\begin{equation}\label{ap:network-kernel}
 K_t\le \frac{2^\nu+\left(2\nu/[e\log(1/\lambda)]\right)^\nu}{1-\lambda}.
\end{equation}
\end{lemma}
\begin{proof}
Set $k=t-1-j$. For $k\le t/2$, $(t/(t-k))^\nu\le2^\nu$ and the geometric sum is bounded by $1/(1-\lambda)$. For larger $k$ bound the factor by $t^\nu$ and the remaining geometric tail by $\lambda^{t/2}/(1-\lambda)$. The maximum of $x^\nu\exp[-x\log(1/\lambda)/2]$ over $x>0$ is $(2\nu/[e\log(1/\lambda)])^\nu$. The $\nu=0$ case is the ordinary geometric sum.
\end{proof}
In the stability calculation of \cite{sun2025towards,sunthesis}, the centered trajectory difference receives local gradient disturbances at each round. Under a harmonic step and growth exponent $a=\mu_0L$, their envelope is proportional to $(j+1)^{a-1}$. Inserting this envelope in Lemma~\ref{lem:mix} at $q=1$ gives a disagreement term proportional to $K_t t^{a-1}$ with $\nu=1-a$. Passing from a Frobenius norm to a sum of client distances costs at most $\sqrt m$. This is the origin of the mixing factor in the source bound. Exact post-update averaging ($A=J$) removes centered rows immediately, so its network penalty is set to zero separately. The finite geometric bound for the forced recursion is not an intrinsic penalty for $A=J$.

\lead{A replace-one calculation for the averaged output} Compare two runs with the same samples and cohorts except for one replaced example at one client. Same-sample local updates increase the distance by at most $1+L\eta$ by smoothness. When the changed example is selected, let $2v$ bound the conditional expected change between the two gradients at the same model. The bound $v=G$ follows from $G$-Lipschitz losses. The smaller choice $v=\sigma_l$ requires the coupled gradient comparison used by the source. A variance bound around unequal conditional means alone would not provide it.

Choose an early cutoff $r\ge1$ and couple the trajectories up to that index. In DFL the altered client is active at each step, hence $\Pr(\text{early hit})\le r/n$. In CFL the client is active with probability $s/m$ and we obtain $\Pr(\text{early hit})\le sr/(mn)$. When the cohort is held for $K$ consecutive steps, take a round boundary for conditioning in the CFL argument below. On no early hit the two states are still identical there. The two early events have different probabilities even when the noise injected in the averaged direction after the cutoff has the same expectation.

For DFL let $d_j$ be the expected average of the client distances at local step $j$, conditional on no hit by $r$. Doubly stochastic mixing cannot increase that average, by the triangle inequality on each mixed row and the unit column sums. After the cutoff the altered client draws the replacement with probability $1/n$, and its weight is $1/m$, giving
\[
 d_{j+1}\le(1+L\eta_j)d_j+2v\eta_j/(mn).
\]
For centralized training, all participating clients begin a round at the same server pair. Conditional on its cohort, unroll this inequality on each member of the round, using an additive $2v\eta_j/n$ only for the affected client. Averaging those trajectories gives it weight $1/s$ when present. Taking the independent cohort expectation multiplies this by $s/m$. This yields exactly the same scalar recursion at round boundaries after unrolling the $K$ steps. This order of conditioning accommodates the shared cohort within a round.

Write $d_N$ for the expected distance of the final averaged output, bounded by the mean distances just derived. With $\eta_j\le\mu_0/(j+1)$, adding $2v/(mnL)$ and multiplying the factors yields
\begin{equation*}\label{ap:net-stable}
 G d_N\le\frac{2vG}{mnL}
   \left\{\prod_{j=r}^{N-1}(1+L\eta_j)-1\right\}
 \le \frac{2vG}{mnL}(N/r)^a .
\end{equation*}
For the product bound, sum $\log(1+L\eta_j)\le a/(j+1)$ and compare to $\int_r^N dx/x$. A different offset in the decaying steps gives the corresponding ratio of shifted indices. Equation~\eqref{ap:net-stable} controls an averaged-model output and also illustrates why another stability analysis can have little explicit graph dependence \cite{bars2023improved}. It gives no consensus estimate for training at individual nodes.

\lead{Proof of the stability comparison in Theorem~\ref{thm:network}} Let $A_0=2vG N^a/(mnL)$, $B_C=sU/(mn)$ and $B_D=U/n$. Apply Lemma~\ref{lem:replace} with \eqref{ap:net-stable}. The centralized envelope is $A_0/r^a+B_Cr$. The network-aware analysis of the client trajectories in \cite{sun2025towards} uses $A_D=A_0(1+6\sqrt m\,\kappa_\lambda)$ in place of $A_0$ and $B_D$ in the early term. Lemma~\ref{lem:kernel} gives an explicit admissible mixing-sum bound for $0<\lambda<1$. For an averaged output, the same conservative envelope follows directly since $A_D\ge A_0$. The coefficient $6$ is the one in the source's trajectory estimate. Equation~\eqref{ap:net-stable} already suffices for the mean-distance argument.

For either pair $(A,B)$, equalizing the two contributions gives $r=(A/B)^{1/(1+a)}$ and envelope $2A^{1/(1+a)}B^{a/(1+a)}$. Using $2\cdot2^{1/(1+a)}\le4$, substituting $v=\sigma_l$ and $N=TK$, and simplifying the powers of $m$ gives \eqref{eq:top-bounds} in the source's balanced-cutoff form. On a specific finite schedule one uses the envelope at an admissible integer cutoff in $[1,N]$, or at a neighboring round boundary for the independent-cohort calculation, and may always cap it by $U$. Rounding and endpoint effects should be retained for a numerical guarantee. Bounded gradients give the corresponding conservative expression with $G$ in place of $\sigma_l$.

At $s=m$ and exact averaging the two balanced expressions agree algebraically, since $m^{a/(1+a)}/m=m^{-1/(1+a)}$. Along a growing graph family with fixed $n,T,K$, the displayed decentralized envelope decays if $\kappa_\lambda=o(\sqrt m)$. With fixed total data $mn$, increasing $m$ also decreases $n$ and the same conclusion does not follow. Finally, the attenuation $\lambda^q$ in \eqref{ap:forced} concerns $q$ exchanges of the same local endpoints. Raising $q$ at fixed bandwidth changes the feasible training schedule, so an accuracy comparison under a fixed number of rounds cannot by itself predict the outcome at a fixed communication budget.